\documentclass[11pt]{article}
\usepackage{arxiv}

\usepackage[utf8]{inputenc}
\usepackage[T1]{fontenc}
\usepackage{mathptmx}  
\usepackage[expansion=false]{microtype}
\usepackage{amsmath,amssymb}
\usepackage{graphicx}
\usepackage{float}
\usepackage{booktabs}
\usepackage{caption}
\usepackage{subcaption}
\usepackage{xcolor}
\usepackage{enumitem}
\usepackage{pifont}
\usepackage[numbers,sort&compress]{natbib}
\usepackage[colorlinks=true,linkcolor=black,citecolor=Maroon,urlcolor=Maroon]{hyperref}
\usepackage{cleveref}

\definecolor{Maroon}{rgb}{0.5,0.0,0.13}
\graphicspath{{figs/}}

\newcommand{\inplace}{\texttt{in\_place}}
\newcommand{\erratum}{\textsc{erratum}}
\newcommand{\fielderr}{\texttt{field+erratum}}
\newcommand{\fieldsel}{\texttt{field+selective}}
\newcommand{\note}[1]{\textsl{#1}}
\newcommand{\kpe}{\texttt{k\_pe}}

\setheadername{Models Take Notes at Prefill: KV Cache Can Be Editable and Composable}

\title{Models Take Notes at Prefill:\\ KV Cache Can Be Editable and Composable}
\author{Bojie Li\\ Pine AI}
\date{\small Code: \url{https://github.com/19PINE-AI/programmable-kv} \\
Website: \url{https://01.me/research/programmable-kv/}}

\begin{document}
\maketitle

\begin{abstract}
Prefix caching reuses prefill only across an exactly shared prefix, so one changed
field invalidates the entire downstream cache.
Yet overwriting the field's own key/value vectors and reusing the rest leaves the model acting on
the \emph{old} value. The reason, established causally across four model families: at prefill the model
has already written the \emph{field-conditioned conclusion} onto downstream
\emph{notes}; the field's own key/value drives under $1\%$ of the decision. Read as a
notebook of memoized conclusions, two capabilities follow.
\textbf{(1) It is editable.} A salient \emph{erratum} amends the notes;
and with \emph{chain-of-thought}, editing the field alone recovers the decision ($1.00$ at
$8$B, ${\sim}1\%$ compute), while without CoT it is ignored.
\textbf{(2) It is composable.} The notes are position-portable, so a precompiled skill can be RoPE-repositioned
and spliced into any context, indistinguishable from full recompute (logit cosine $0.90$--$0.999$, twelve
models) at $O(L)$ rather than $O(L^2)$ time-to-first-token. A unified edit+compose agent stays
decision-identical to recompute at up to $14.9\times$ lower latency.
The approach applies to any per-token attention KV cache, validated across scale, quantization,
Mixture-of-Experts, and multimodal caches, and extends to several attention variants through small
adapters. Because the erratum is append-only, it composes with production prefix caching: in an online
vLLM benchmark it keeps the prefix cache-aligned ($98.5\%$ hit-rate), cutting $p90$ time-to-first-token by
$53$--$398\times$.

\end{abstract}

\section{Introduction}
\label{sec:intro}

Modern LLM agents re-read long, mostly-static instructions on every turn---a system policy, a tool
specification, retrieved documents. Key/value (KV) caching makes this affordable by reusing the prefill
across turns, but only across an \emph{exact} shared prefix. The moment one token changes inside the reused
region---a timestamp, a user id, an order's status---the keys and values of \emph{every} later token are
invalidated, because each attended to the token that changed. The de-facto workaround---\emph{hoisting} all
mutable content to the end so the static prefix stays cache-aligned---pushes an inference-layer constraint
into the application layer: fields referenced in several places, nested sub-agent prompts, and dynamically
assembled contexts cannot all be cleanly hoisted, and the application must enumerate every mutable field in
advance.

\begin{figure}[tbp]
\centering
\includegraphics[width=\linewidth]{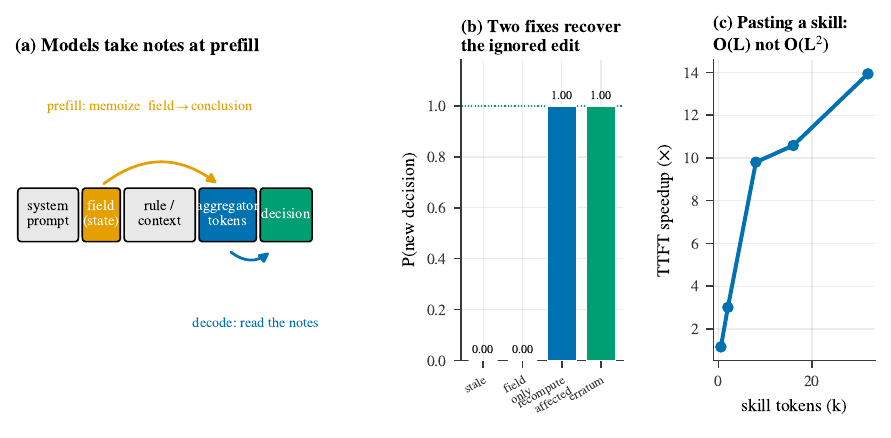}
\caption{\textbf{Models take notes at prefill.} (a) At prefill the model memoizes the field-conditioned
conclusion onto downstream aggregator tokens (orange); at decode the decision reads those notes (blue).
(b) Consequently, surgically editing the field's own KV is ignored (without a reasoning chain), but the
decision is recovered cheaply: recompute the \emph{affected} downstream suffix, or---cheaper and robust---
append a salient \emph{erratum} (P(new decision), no-CoT model; with chain-of-thought the field-only
refresh alone also works, \Cref{fig:overview}c). (c) Because the notes are position-portable, a precompiled
skill can be pasted into a new context in $O(L)$ rather than $O(L^2)$ time.}
\label{fig:teaser}
\end{figure}

This paper starts from a concrete puzzle. The region of the cache \emph{before} a field is, by
construction, independent of the field's value---we measure a key/value deviation of exactly $0.0$ when
the field changes. One might therefore hope to \emph{surgically} refresh only the field's own keys and
values, leave the rest of the cache stale, and pay almost nothing. We find this fails completely: the
model's decision reverts to the \emph{old} field value, as if the edit never happened
(\Cref{fig:teaser}b).

\paragraph{The discovery.} The reason, which we establish causally, is that transformers do not defer
their reasoning to decode time. At \emph{prefill} the model already computes the
\emph{field-conditioned conclusion} and writes it onto downstream tokens---disproportionately onto
aggregator/delimiter tokens that later positions attend through. The decision then reads these
\emph{notes}, not the field itself: across models the field's own KV causally drives less than $1\%$ of
the decision, while the downstream notes drive essentially all of it. The KV cache is best understood not
as a frozen byproduct of prefill but as a \emph{notebook of memoized conclusions} (\Cref{fig:teaser}a).

\begin{figure}[tbp]
\centering
\includegraphics[width=\linewidth]{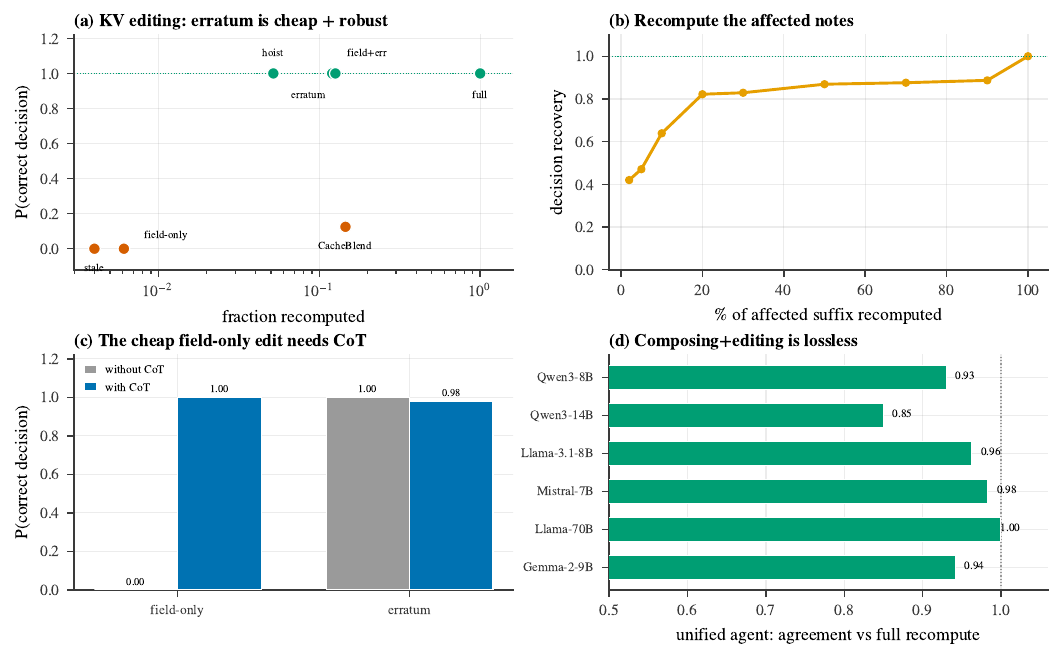}
\caption{\textbf{Editing and composing, previewed} (Qwen3-8B unless noted; detail in
\Cref{sec:mechanism,sec:editable,sec:keystone}). (a) \emph{KV editing landscape}: naive edits (stale,
field-only without CoT, CacheBlend) fail, while the append-only \texttt{erratum}/\texttt{field+erratum}
reach full-reprefill correctness cheaply and robustly (hoisting also works but needs prompt surgery).
(b) \emph{Recompute the affected notes}: recovery climbs as more of the post-field \emph{affected suffix} is
recomputed---reliable in the limit, but the cheap top-$K$ version (\texttt{field+selective}@$K$) is
unreliable (\Cref{fig:edit}). (c) \emph{Chain-of-thought, not model size, gates the cheapest edit}: with CoT
the near-free field-only refresh alone recovers the decision ($1.00$); without CoT it is ignored ($0.00$);
the erratum fixes both. (d) \emph{Composing $+$ editing is lossless}: the unified agent stays
decision-identical to full recompute across models.}
\label{fig:overview}
\end{figure}

\paragraph{Two capabilities from one mechanism.} Once the cache is read as a notebook of memoized
conclusions, we can manipulate those conclusions directly---editing or reusing them in place rather than
recomputing them---in two ways.
(1) If the conclusion is
already written downstream, then \emph{editing} a field means amending the notes, not recomputing them. The
cheap and robust fix is a one-line salient \emph{erratum} that overrides the stale notes; one can instead
recompute the affected downstream notes (reliably for the full affected suffix, cheaply-but-unreliably for
a top-$K$, \fieldsel{}@$K$). An even cheaper in-place refresh of the field alone is gated by
\emph{chain-of-thought}: with a reasoning chain the model re-reads the field and the edit suffices, without
it the edit is ignored. (2) If the notes are localized and position-portable, then a reusable skill can be
\emph{composed} into a new context by repositioning and splicing its cached notes---no recompute. A
\emph{unifying} experiment, editing a field \emph{inside} a transplanted skill, shows the two operations
act on the same notes. \Cref{fig:overview} previews both capabilities. More broadly, we view editing and
composition as first instances of a \emph{programmable} KV cache---a structured memory that systems, and
eventually models trained to expose it, can read, write, and rearrange rather than only extend linearly.

\paragraph{Structure of the paper.}
\begin{itemize}[leftmargin=1.2em,itemsep=1pt,topsep=2pt]
\item \textbf{A mechanism} (\Cref{sec:mechanism})---\emph{attention-mediated memoized inference}---established
by four causal probes (locality patching, suffix concentration, linear probing, circuit knockout) and four
further controls (content/conclusion dissociation, layer-timing, specificity, note-injection;
\Cref{app:deepmech}), replicated across \emph{four model families} (Qwen3, Llama-3.1, Gemma-2, Mistral),
resolved to a \emph{component-level circuit}---named read/write heads, a causal conclusion direction, an SAE
feature, attention-vs-MLP, and causal scrubbing (\Cref{app:circuit})---and connected to the delimiter-token
aggregation seen in interpretability.
\item \textbf{An editing capability} (\Cref{sec:editable}): naive KV editing fails; the erratum /
\texttt{field+erratum} fix matches the hoist-to-end oracle without prompt surgery; an analysis of when
the ${\sim}1\%$-compute in-place edit suffices (it requires reasoning and is strongly
model-dependent); and a head-to-head with \emph{weight}
editing (ROME, LoRA) showing it is the wrong tool for mutable per-request state (global contamination,
collateral, $30$--$50\times$ slower).
\item \textbf{A composing capability} (\Cref{sec:composable}): position-portable transplant of precompiled
skills, $O(L)$ vs.\ $O(L^2)$ TTFT ($13.9\times$ at $32$k), with a seam-repair knob---building on prior
caching work (\Cref{sec:related}), our addition being the mechanism that explains it and a correctness lens.
\item \textbf{The unification} (\Cref{sec:keystone}): this experiment and a unified edit+compose agent over
thirteen models.
\item \textbf{An application: user memory} (\Cref{sec:memory}): the large, mutable user-memory
document is both composed (precompiled, repositioned, spliced) and edited (in-place / erratum)
as one set of notes---decision-faithful to full recompute at $2.3$--$4.3\times$ lower
time-to-first-token, validated to $70$B and on real long-conversation memory (LoCoMo, transplant
$\equiv$ full recompute in QA accuracy)---with a pre-registered, statistically-controlled evaluation.
\item \textbf{Applicability to multimodal and new attention mechanisms} (\Cref{sec:reach}): the mechanism carries from small models through MoE and
low-bit quantization to multimodal image caches; lightweight adapters bring it to MLA and interleaved
M-RoPE; a mask-based remedy handles sliding-window attention; and we chart where it breaks down against the
2026 sparse/compressed-attention frontier.
\item \textbf{Systems payoff} (\Cref{sec:systems}): a real agentic environment and a comprehensive online
vLLM serving benchmark (V1 engine, continuous batching, Poisson load)---$98.5\%$ vs.\ $1\%$ prefix-cache
hit-rate, $53$--$398\times$ lower $p90$ TTFT, and throughput gains that grow with load to $14.5\times$.
\end{itemize}

\section{Related work}
\label{sec:related}

\paragraph{Where computation is stored, and how to edit it.} A line of interpretability work localizes
stored \emph{knowledge} in transformer weights and edits it there: ROME and MEMIT \citep{meng2022rome,
meng2023memit} locate and rewrite factual associations in MLP weights, while circuit analyses such as the
indirect-object-identification study \citep{wang2023ioi} trace how specific computations are carried by
attention heads. Weight editing targets \emph{durable, global} facts; we compare against a faithful ROME
and a LoRA fine-tune empirically (\Cref{tab:weight}) and find them ill-suited to \emph{mutable per-request}
state---a global edit contaminates concurrent requests and damages unrelated decisions---which is exactly
the niche the editable KV cache fills. Closest in spirit, \citet{lindsey2025biology} find models commit \emph{plans} to
specific tokens (e.g.\ a planned rhyme stored on a line-break token) during the forward pass. We study a
complementary object: not weights and not decode-time computation, but the \emph{KV cache}---the
activations an inference system already stores and an editor can directly manipulate---and show it holds
\emph{memoized conclusions} concentrated on aggregator/delimiter tokens. To our knowledge this is the
first causal account of why in-place KV field editing fails and what to do instead.

\paragraph{KV reuse and composable caching.} Reusing precomputed KV beyond an exact prefix is an active
systems topic, and our \emph{composing} capability builds directly on it; we claim none of the caching
machinery as novel. Prompt Cache \citep{gim2024promptcache} precomputes reusable prompt modules with
position placeholders and splices them. CacheBlend \citep{yao2025cacheblend} reuses non-prefix chunk KV
and selectively recomputes ${\sim}15\%$ of tokens to restore cross-attention. EPIC
\citep{hu2025epic} introduces position-independent caching with \textsc{AttnLink}, recomputing only a few
chunk-boundary tokens (exploiting attention sinks) for near-linear recompute. CacheSlide
\citep{liu2026cacheslide} reuses KV in a position-\emph{aware} way via relative-position-dependent caching,
and MPIC \citep{mpic2025} extends position-independent caching to the multimodal setting by recomputing
image-boundary tokens; KVLink \citep{yang2025kvlink} is a further reuse system. Mapped onto our terms, our
RoPE-repositioning is CacheSlide's relative-position reuse, our seam-repair is the boundary recompute of
CacheBlend/EPIC/MPIC, and our image-KV transplant is MPIC's idea (we \emph{re-rotate} M-RoPE rather than
recompute). Our contributions over this line are orthogonal: (i) the \emph{mechanism} that explains
\emph{why} boundary recompute is what is needed; (ii) a \emph{decision-governance} evaluation---does a
transplanted skill still \emph{govern the tool decision}, rather than only preserve perplexity or
throughput; (iii) the \emph{editing} axis and the edit+compose unification; and (iv) adapters that extend the
operations to new attention representations (MLA, interleaved M-RoPE, sliding-window). We note honestly
that this editable/composable-cache direction grew directly out of the position-independent and
position-aware caching of EPIC and CacheSlide \citep{hu2025epic,liu2026cacheslide}, and out of discussions
with Junhao Hu (first author of EPIC; see Acknowledgements).

\paragraph{Prefix caching, KV compression, and reuse systems.} Production prefix caching (vLLM Automatic
Prefix Caching, SGLang RadixAttention) reuses exact prefixes. A large literature instead \emph{compresses}
or \emph{evicts} the cache: StreamingLLM keeps attention sinks \citep{xiao2024streamingllm}; H2O
\citep{zhang2023h2o}, Scissorhands \citep{liu2023scissorhands}, and SnapKV \citep{li2024snapkv} evict
low-importance tokens; Quest \citep{tang2024quest} keeps all tokens but attends sparsely. Serving systems
stream or share cached KV across requests---CacheGen \citep{liu2024cachegen} for fast loading and RAGCache
\citep{jin2024ragcache} for retrieval reuse. These change \emph{which} tokens are present, and compose with
our edit/transplant operations only over retained tokens; we treat them, the latent/decoupled-RoPE
representation of Multi-head Latent Attention \citep{deepseekv2,deepseekv3}, and the 2026
sparse/compressed-attention designs as scope boundaries in \Cref{sec:reach}. Our repositioning relies on
rotary position embeddings \citep{su2024rope} and their extensions \citep{peng2024yarn}.

\paragraph{Activation-level interventions.} Beyond weight editing, a body of work intervenes on
\emph{activations}: steering and inference-time interventions \citep{li2023iti}, task and function vectors
\citep{ilharco2023taskvectors,todd2024functionvectors}, and the causal-mediation/activation-patching
methodology we adapt \citep{vig2020causal}. These edit residual-stream directions to change behavior; we
instead read and write the \emph{KV cache} itself---the per-token activations a serving system already
persists---which is what makes the intervention both interpretable and deployable.

\paragraph{Agents and tool use.} Our evaluation targets tool-using agents in the style of ReAct
\citep{yao2023react} and Toolformer \citep{schick2023toolformer}, and we measure end-to-end task success on
the $\tau^2$-bench tool-agent-user benchmark \citep{barres2025tau2}, the dual-control successor of
$\tau$-bench \citep{yao2024taubench}, where state changes mid-trajectory make
cache staleness a first-class correctness problem.

\section{The discovery: memoized inference in the KV cache}
\label{sec:mechanism}

\begin{figure}[tbp]
\centering
\includegraphics[width=\linewidth]{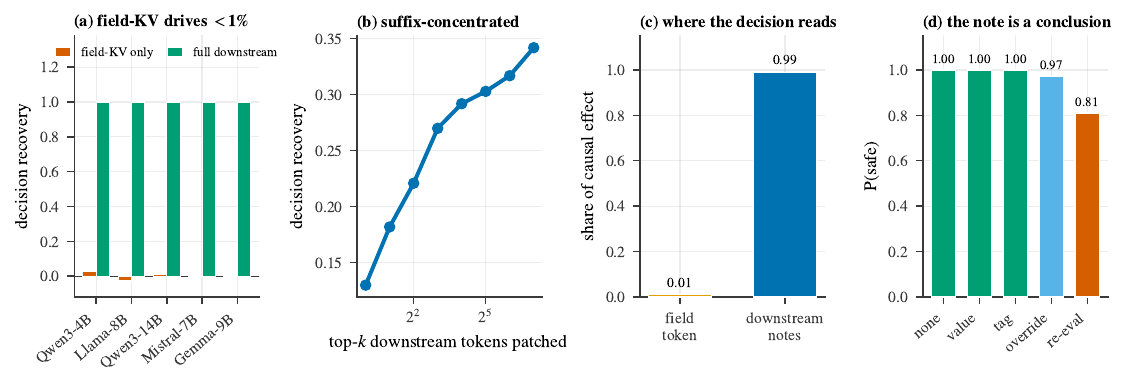}
\caption{\textbf{Four causal probes for memoized inference.} (a) Refreshing the field's own KV recovers
$\approx 0$ of the decision; recomputing the downstream recovers it fully. (b) Recovery is
suffix-concentrated, accruing only as many post-field tokens are patched. (c) The decision reads almost
entirely from downstream notes, not the field token. (d) Once the value is present, override wording is
redundant and ``re-evaluate'' phrasing hurts---the note is a committed conclusion.}
\label{fig:mech}
\end{figure}

\paragraph{Where the conclusion is written and read.} We first state the account that the probes in this
section establish. A transformer prefills the prompt left-to-right, each position attending to all earlier
ones. At a small number of positions \emph{after} the mutable field---\emph{aggregator} tokens such as
punctuation, newlines, and section breaks, which later tokens route attention through---the model does more
than cache the raw token: it computes the field-conditioned answer (e.g.\ ``status is \texttt{shipped}, so
the action is \emph{deny}'') and writes that conclusion into those positions' cached key/value vectors at
mid-to-late layers. These cached vectors are the \emph{notes}, and they lie \emph{downstream} of the field.
At decode time the decision token does not re-derive the answer from the field; it attends back to the note
positions, and a small set of late-layer attention heads read the stored conclusion into the output logit.
Refreshing the field's \emph{own} KV therefore changes little, since the conclusion was written elsewhere and
the field itself accounts for under $1\%$ of the decision. The four probes below establish each link of this
chain causally: \emph{where} the notes lie, that they encode a \emph{conclusion} rather than a copy of the
field, and that the decision \emph{reads} them.

\paragraph{Setup.} We study a minimal but agent-realistic decision: a context contains a policy rule and a
\emph{mutable field} whose value determines the correct action (e.g.\ ``cancel an order only if its status
is \texttt{pending}''; with \texttt{status=shipped} the correct action flips from \emph{cancel} to
\emph{deny}). After prefilling the full context we compare four cache states at the decision token:
\emph{stale} (old field, old downstream), \emph{field-only} (refresh the field's KV, leave the downstream
stale), \emph{full-downstream} (recompute everything after the field), and \emph{oracle} (a clean prefill
of the new value). We report \emph{decision recovery}: the fraction of the oracle's flip that a cache
state reproduces ($0$ = behaves like stale, $1$ = behaves like oracle). Four independent probes converge
on one account (\Cref{fig:mech}); \Cref{app:examples} gives a complete worked example---the verbatim
prompt, the erratum, and recorded model responses.

\paragraph{(1) Locality: the field's own KV barely matters.} Refreshing only the field's KV recovers
essentially none of the decision---field-only recovery is $-0.028$ on Llama-3.1-8B and near zero across
models---whereas recomputing the downstream recovers it fully ($1.0$) (\Cref{fig:mech}a). The field is
read \emph{indirectly}: its causal contribution to the decision is under $1\%$.

\paragraph{(2) Suffix concentration: the effect lives downstream and late.} Sweeping how many downstream
tokens we patch (ranked by causal effect) shows recovery accrues slowly and saturates only as we include
many tokens after the field, with the causal mass concentrated in mid/late layers
(\Cref{fig:mech}b). The information the decision needs is not in the field but distributed over the tokens
that followed it at prefill (\Cref{fig:mech}c).

\paragraph{(3) Linear decodability.} The field-conditioned conclusion is linearly decodable from those
downstream tokens' residual stream at prefill time---i.e.\ the model has already \emph{computed and
written down} the answer, not merely copied the field.

\paragraph{(4) Knockout and dose-response.} Ablating the high-effect downstream tokens flips the decision
back to stale, and the effect grows with the number/position of memoizing tokens, ruling out a diffuse
explanation: specific aggregator/delimiter positions carry the conclusion. This mirrors the delimiter-token
aggregation reported by \citet{lindsey2025biology} for forward planning; here the stored quantity is a
backward-looking, field-conditioned \emph{conclusion}.

\paragraph{What the note contains.} If the note were a verbatim copy of the field, restating the value
would suffice and emphatic wording would be inert. Instead, a wording ablation (\Cref{fig:mech}d) shows
that once the corrected value is present, the override phrasing is \emph{redundant} (bare value, tagged
update, and explicit override all reach $\approx 1.0$ P(safe)), while aggressive ``disregard your earlier
conclusion and re-evaluate'' phrasing actively \emph{hurts} ($0.81$). The note behaves like a committed
conclusion that a late, salient correction can overwrite---but that confrontational instructions can
destabilize. We name the phenomenon \textbf{attention-mediated memoized inference}. The remainder of this
section stress-tests the account---further controls, more architecture families, and a component-level
circuit---before the rest of the paper builds on it.

\paragraph{Separating the stored conclusion from the field content.} A linear probe can \emph{decode} the
conclusion from the downstream tokens, but decodability alone does not show the decision \emph{uses} it---the
same tokens also encode the field's raw content, and a probe finds both. Four causal controls (three models
each: Qwen3-8B/4B and Llama-3.1-8B; \Cref{app:deepmech}) close that gap; the causal patch, not the probe, is
the instrument that separates the two.

\textbf{(i) It is the conclusion, not the field content.} We hold the field value byte-identical and flip a
single rule token (a polarity \emph{trigger}) so the \emph{conclusion} inverts while the field \emph{content}
stays constant. Transplanting the downstream notes then carries the entire flipped conclusion (recovery
$0.998$--$1.009$), whereas patching the changed rule token carries none ($-0.007$ to $+0.007$). Since the
content never changed, the notes cannot merely be re-encoding it.

\textbf{(ii) The note is written before it is read.} Within the single prefill pass the conclusion becomes
decodable on the downstream aggregator about twelve layers \emph{earlier} (relative depth $0.31$--$0.39$)
than the point at which the decision token commits to its answer (depth $0.73$--$0.77$): the note already
exists at prefill, ahead of the read.

\textbf{(iii) A few specific tokens carry it.} Transplanting the eight \emph{highest-effect} downstream
positions recovers $0.74$--$0.79$ of the decision, while eight \emph{random} downstream positions recover
${\le}0.035$: the conclusion lives in a few specific aggregator tokens, not a diffuse code.

\textbf{(iv) The decision follows the note, even when the note is wrong.} Injecting a \emph{false} note (the
opposite conclusion's downstream KV) into an otherwise-consistent cache makes the decision follow the written
note \emph{against its own live field} (recovery $\approx1.0$); a handful of note tokens suffice. The account
also holds off the synthetic template: field-only recovery stays ${\approx}0$ for multi-hop reasoning and
free-form conversational phrasing, and is bounded only for near-verbatim attribute lookup, where the field is
itself partly a copy (\Cref{app:deepmech}).

\paragraph{Not a model-family artifact: four families.} To rule out the aggregator-token account being
a Qwen3/Llama tokenizer artifact, we replicate five probes (the locality probe and the four deep
controls) on two further architecture families,
\textbf{Gemma-2-9B} and \textbf{Mistral-7B} (a tokenizer-robust readout; for Gemma-2 we keep its
attention/logit soft-capping intact). Every result holds: field-only recovery $0.005$/$0.137$ vs.\
full-downstream $1.0$; the conclusion/content dissociation (trigger-only $\approx0$ with the field held
identical vs.\ notes $\approx1.0$); top-$8$ vs.\ random-$8$ specificity ($0.95$ vs.\ $0.48$); false-note
injection ($0.98$--$1.0$, follow-rate $1.0$); and write-before-read timing (write depth $0.19$--$0.26$ vs.\
decision commit $0.47$--$0.48$). The mechanism is consistent across \emph{four} families (Qwen3, Llama-3.1,
Gemma-2, Mistral; \Cref{app:deepmech}).

\paragraph{A component-level circuit.} The probes above show \emph{where} the conclusion is stored; five
further interventions (\Cref{app:circuit}; replicated across Llama-3.1, Qwen3, Gemma-2, and Mistral) show
\emph{which components} write and read it. The pattern is \emph{distributed write, concentrated read}.

\textbf{Write (many components, mid layers).} Mid-layer \emph{attention}---not the MLPs---does most of the
writing onto the aggregator tokens ($\approx0.6$ of the write on Llama). The conclusion is stored
\emph{redundantly}: it is easy to \emph{decode} (a trained sparse-autoencoder feature separates the two
conclusions at AUC $1.0$), yet its causal content is spread across ${\sim}10$--$30$ features along a shared
low-rank direction (that direction carries $\approx25\times$ the decision-effect of a random one).

\textbf{Read (few components, late layers).} A small, nameable set of late \emph{read heads}---those
attending from the decision token back to the aggregators---funnels the stored conclusion into the output
logit ($12$ such heads recover $0.78$ of the decision; the same number of random heads recover $\approx0$).

\textbf{Control.} Causal scrubbing confirms the note alone governs the decision: resampling it to the
opposite conclusion flips the decision, while resampling everything else leaves it unchanged. As at the
token level, the single feature that best \emph{decodes} the conclusion is not by itself \emph{causally}
sufficient---decodability and causation come apart, now among individual features. With the mechanism
established, the rest of the paper turns it into capabilities: the cache is editable (\Cref{sec:editable})
and composable (\Cref{sec:composable}).

\section{Consequence I: the cache is editable}
\label{sec:editable}

\begin{figure}[tbp]
\centering
\includegraphics[width=\linewidth]{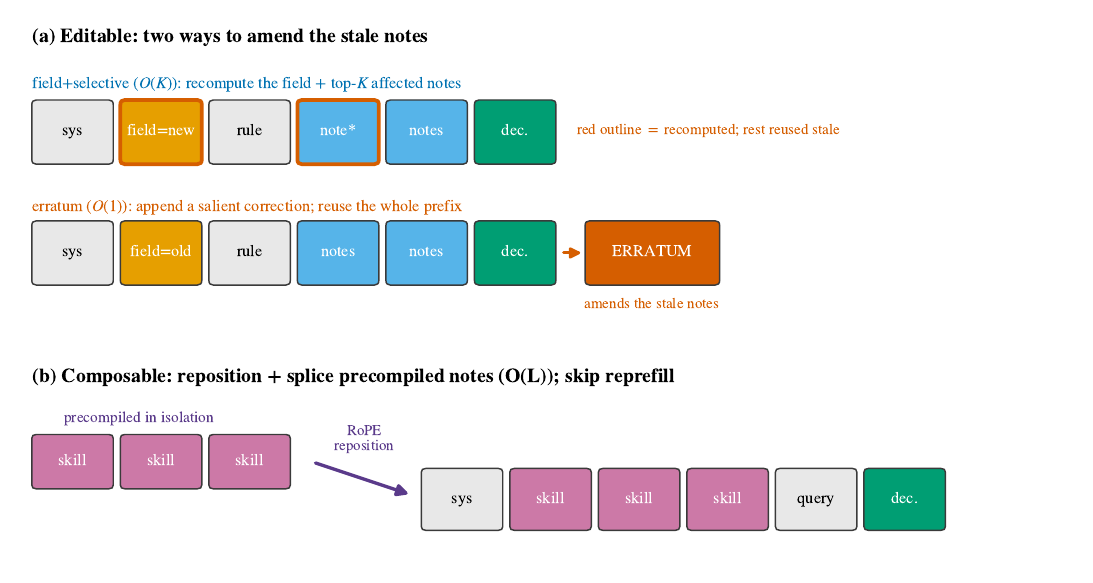}
\caption{\textbf{Two cache operations.} (a) \emph{Editable}, two ways to amend the stale notes:
\fieldsel{} recomputes the field plus the top-$K$ highest-effect downstream notes ($O(K)$; recomputed cells
outlined in red, the rest reused stale), while the \texttt{erratum} appends one salient correction ($O(1)$,
the whole prefix is reused). (b) \emph{Composable}: precompute a skill's KV in isolation, RoPE-reposition it
to the target positions, and splice it in ($O(L)$, no reprefill).}
\label{fig:ops}
\end{figure}

The mechanism tells us how to repair the cache after a field changes, not only why a naive edit fails.
Because the decision reads a conclusion stored in the downstream notes (\Cref{sec:mechanism}),
\emph{editing} means amending those notes rather than recomputing the prefix; this section establishes when a
naive edit fails, the two ways to repair the notes, and which to prefer. \Cref{fig:ops} contrasts the two cache operations this section and the
next make precise: editing appends a salient correction; composing repositions and splices precompiled KV.

\paragraph{Naive editing fails---by design.} The mechanism predicts the in-place edit will fail, and it
does: refreshing the field's KV while reusing the stale downstream leaves the decision at the old value
(\Cref{fig:teaser}b). The prefix before the field is genuinely reusable (deviation $0.0$); the problem is
that the conclusion was already memoized after it.

\paragraph{Two ways to fix it, neither a full reprefill.} If the stale notes carry an old conclusion, there
are two interventions. \emph{(i) Recompute the affected notes} after the field. The safe version recomputes
the \emph{whole} suffix after the field---exactly the \emph{full-downstream} state from \Cref{sec:mechanism}
(recovery $1.0$)---but pays the post-field prefill. The cheaper version, \fieldsel{}@$K$, recomputes only the
field plus the $K$ downstream tokens that carry most of the decision---the high-effect aggregator positions
identified in \Cref{sec:mechanism}, ranked by their causal effect on the decision---and reuses every other
cached token unchanged. It is cheaper but \emph{unreliable}: the smallest $K$ that restores the decision is
strongly model-dependent (below). \emph{(ii) Append an erratum}:
one salient line late in the context---``\texttt{[STATE UPDATE] field $\rightarrow$ new; overrides any
earlier value and conclusion}''---so the decision token attends to a fresh, authoritative note (verbatim
template in \Cref{app:examples}). The erratum is the cheap, robust default: appended after the field
(\texttt{field+erratum}) it matches the strong \emph{hoist-to-end} oracle \emph{without rewriting the
prompt} (P(correct) $1.00$ on the gated-task frontier, \Cref{fig:overview}a; \Cref{tab:frontier}), and being
append-only it composes with prefix caching (\Cref{sec:systems}).

\begin{figure}[tbp]
\centering
\includegraphics[width=\linewidth]{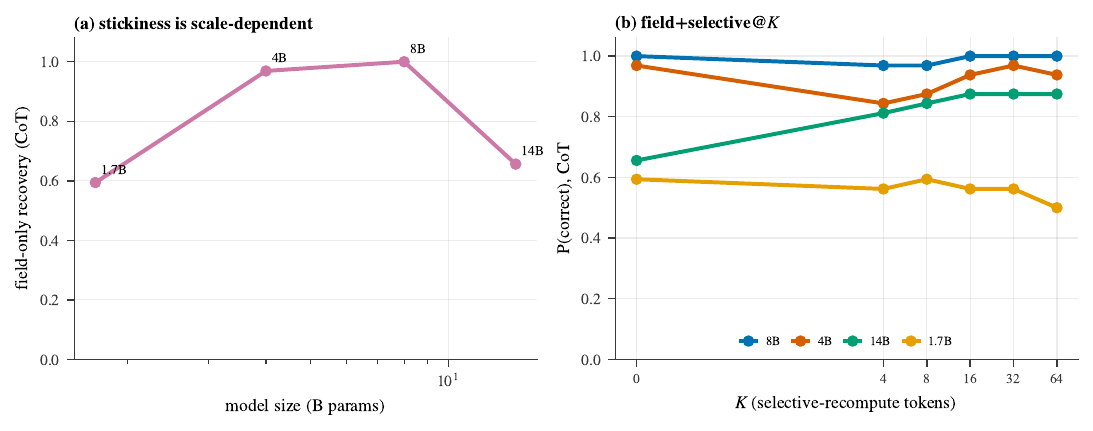}
\caption{\textbf{Editing is model-dependent} (the editing landscape and the chain-of-thought split are
previewed in \Cref{fig:overview}). (a) \emph{Given} chain-of-thought (which gates whether the edit works at all), the
residual stickiness is model-dependent: the near-free field-only recovery varies non-monotonically across
model sizes. (b)
\texttt{field+selective}@$K$: the minimal recompute $K^\star$ to reach full quality is likewise
model-dependent.}
\label{fig:edit}
\end{figure}

\paragraph{Chain-of-thought, not model size, gates the cheapest edit.} A third, near-free option---refresh
the field's own KV alone (${\sim}1\%$ compute) and nothing else---works only when some later computation
actually re-reads the field. A reasoning \emph{chain} does exactly that, so with chain-of-thought the
field-only edit alone recovers the decision ($1.00$ on Qwen3-8B); without the chain the \emph{identical}
edit on the \emph{same} model is ignored ($0.00$) and the decision commits to the stale note
(\Cref{fig:overview}c). The divider is therefore the CoT \emph{mode}, not raw scale---reasoning-native
models default to CoT, instruction-tuned ones to direct answers---so without CoT one of the two real fixes
above is required. The dependence is two-level, and it is worth stating precisely: CoT gates \emph{whether}
the field-only edit can work at all (the binary $0.00$/$1.00$ split above is set by mode, not size), while
\emph{conditional on} CoT the residual \emph{stickiness}---how completely the cheap edit recovers the
decision---is model-dependent (\Cref{fig:edit}a; and E3 of \Cref{sec:memory}, where it climbs with scale
within the Qwen3 family). These two dependencies---CoT-gating and per-model stickiness---are the whole story for the cheap edit, and we keep the rest brief here; the per-model
stickiness, the \fieldsel{}@$K$ sweep ($K^\star\approx 4$ at $8$B but $>64$ at $4$B, varying
non-monotonically with scale; \Cref{fig:edit}a,b), and the layer-wise account are detailed in
\Cref{app:edit,app:deepmech}. We report
\fieldsel{}@$K$ honestly as a genuine but \emph{unreliable} surgical tool, not a default.

\paragraph{Which edit to use: no single dominant method.} A controlled comparison (full table in \Cref{app:edit}) shows no single
dominant method. Hoist-to-end is cheapest but demands prompt surgery and pre-identification of every field;
\texttt{field+erratum} matches it with no surgery at a one-line append; \inplace{} is near-free but only
with chain-of-thought; a KV-deviation-ranked selective recompute (CacheBlend-style \citep{yao2025cacheblend}) underperforms here
because it chases changed keys rather than the tokens that \emph{memoized the conclusion}. The practical
recommendation is \texttt{field+erratum} as the robust default, with \inplace{} as a free fast-path under
chain-of-thought.

\paragraph{Why not edit the weights?} A natural objection: to act on a changed field, why not edit
the model's \emph{weights} (ROME/MEMIT \citep{meng2022rome,meng2023memit}) or fine-tune, rather than the cache? We compare on the paper's
gated task (status \texttt{pending}$\rightarrow$\texttt{shipped}, so \emph{cancel}$\rightarrow$\emph{deny})
against a faithful rank-one ROME---validated on the canonical factual edit first (``the Eiffel Tower is
in'' \emph{Paris}$\rightarrow$\emph{Rome}, locality intact), so the baseline is not crippled---and a LoRA
fine-tune \citep{hu2022lora} (\Cref{tab:weight}; methodology in \Cref{app:systems}). All of \fielderr{}, ROME, and LoRA \emph{succeed} at flipping the target
decision. But a weight edit is \emph{global}: the same model instance can no longer hold
\texttt{status=shipped} for one request and \texttt{pending} for another, so \emph{all} concurrent orders
that are genuinely still pending are wrongly flipped (cross-request contamination $1.0$), and half of an
unrelated decision battery drifts ($0.5$)---the ROME/fine-tune specificity tax---at $3$--$6$\,s per edit
(plus a one-time covariance pass for ROME). The append-only erratum lives in a \emph{per-sequence} cache:
zero cross-request contamination, zero collateral, $114$\,ms, and it composes with prefix caching
(\Cref{sec:systems}). Weight editing targets \emph{durable, global facts}; mutable per-request, per-turn
state is the wrong job for it---which is precisely the niche the editable KV cache fills.

\begin{table}[tbp]
\centering\small
\caption{\textbf{KV editing vs.\ weight editing} for mutable per-request state (Llama-3.1-8B). Weight edits
succeed at the target yet are global (contaminate concurrent requests), damage unrelated decisions, and are
$30$--$50\times$ slower per edit.}
\label{tab:weight}
\begin{tabular}{lcccc}
\toprule
method & flips decision? & edit latency & cross-req.\ contamination & collateral \\
\midrule
\fielderr{} (KV) & \ding{51} & $114$\,ms & $\mathbf{0}$ & $\mathbf{0}$ \\
\inplace{} (KV) & \ding{55}\,(no CoT) & $71$\,ms & $0$ & $0$ \\
ROME (rank-one) & \ding{51} & $5.6$\,s\,$+\,11.6$\,s cov & $1.0$ & $0.5$ \\
LoRA fine-tune & \ding{51} & $3.1$\,s & $1.0$ & $0.5$ \\
\bottomrule
\end{tabular}
\end{table}

\section{Consequence II: the cache is composable}
\label{sec:composable}

\begin{figure}[tbp]
\centering
\includegraphics[width=\linewidth]{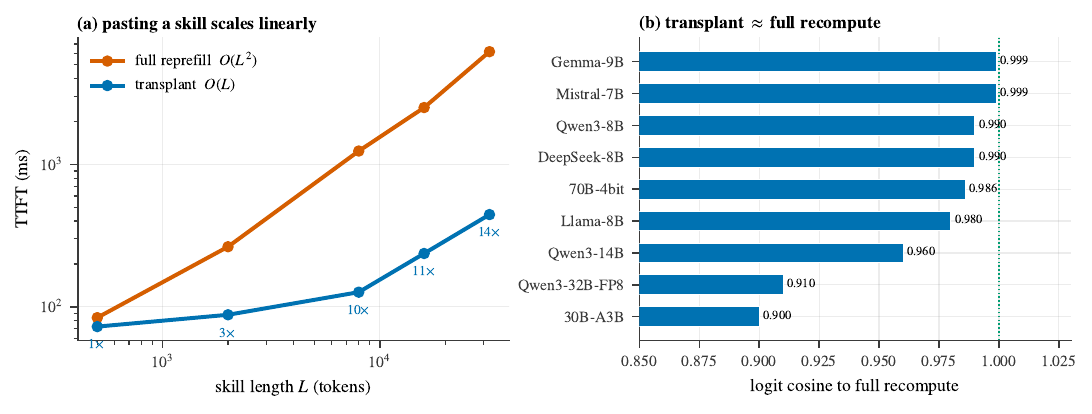}
\caption{\textbf{Composing the cache.} (a) Pasting a precompiled skill is $O(L)$ vs.\ full reprefill's
$O(L^2)$; TTFT speedup reaches $13.9\times$ at $32$k tokens. (b) The transplanted skill matches full
recompute in next-token logits (cosine $0.90$--$0.999$) across the model family.}
\label{fig:comp}
\end{figure}

\paragraph{From mechanism to prediction.} If a region's notes are localized and re-derivable from context
the decision can still see, then the notes should be \emph{position-portable}: we can compute a reusable
chunk's KV once, in isolation, move it to a new absolute position, and splice it in. Concretely we
precompile a long \emph{skill} (a policy or tool specification), and because the attention library caches
post-RoPE keys, we re-rotate the chunk's keys from their source positions to the target positions (values
are position-free) before concatenating (\Cref{fig:ops}b). This is the position-aware reuse of
\citet{liu2026cacheslide}; our point is that the \emph{mechanism predicts the splice should match a full
reprefill}---the same next-token logits and decisions.

\paragraph{Transplant is behaviorally indistinguishable from full recompute.} It does. The spliced skill
matches a full reprefill in next-token logits with cosine $0.90$--$0.999$ across the full model family
(twelve models; full roster in \Cref{app:models})---Qwen3-1.7B through 32B (including FP8 and the 30B-A3B
Mixture-of-Experts), Gemma-2/3, Mistral-7B, Llama-3.1-8B and 70B, and DeepSeek-R1-Distill-Llama-8B
(\Cref{fig:comp}b)---and on the models competent at the task it
preserves \emph{correct} skill-following across $8$ diverse domains and $3$ families ($24/24$, cosine
$0.98$--$0.999$), including $16/16$ under chain-of-thought.

\paragraph{Context-robustness and the seam.} A chunk precompiled in isolation matches one that attended to
the real preceding context, because the decision re-derives from context it still sees. The one residual
error is a \emph{seam} at the chunk's start---the first tokens that, in a full prefill, would have
attended to the now-missing prefix. Recomputing a few boundary tokens (\emph{seam-repair}) closes it; this
is exactly the boundary recompute of CacheBlend/EPIC/MPIC, and \Cref{sec:mechanism} explains why it is
the boundary, specifically, that needs repair.

\paragraph{Linear-time TTFT.} Full reprefill of a length-$L$ skill is $O(L^2)$; transplant is $O(L)$ (a
re-rotation pass plus the suffix). Time-to-first-token speedups grow with skill length: $3\times$ at $2$k
tokens, $9.8\times$ at $8$k, and $13.9\times$ at $32$k on an $8$B model (\Cref{fig:comp}a). A library of
skills composes (decisions preserved for $N=1$--$4$ concurrent skills).

\paragraph{Generality: content type, insertion position, and agentic tool-calling.} Transplantation is not specific to rule-like
skills. It preserves decisions for \emph{facts/RAG} passages as well as rules; for chunks inserted in the
system area \emph{and} mid-trajectory as tool results; and---measured with actual function calls rather
than a proxy---it preserves \emph{agentic tool-calling} ($N{=}108$ with bootstrap CIs: function-call
accuracy $1.00$ on Mistral-7B, Llama-3.1-8B/70B, and Qwen3-8B; $0.97\,[0.94,1.0]$ on Qwen3-32B-FP8 and
30B-A3B, whose transplanted-vs-full tool-call \emph{agreement} is $1.00$). The one consistent exception
is sliding-window attention (Gemma), which we diagnose and fix in \Cref{sec:reach}. Full per-domain
scorecards are in \Cref{app:compose}.

\section{Combining edit and compose}
\label{sec:keystone}

\begin{figure}[tbp]
\centering
\includegraphics[width=\linewidth]{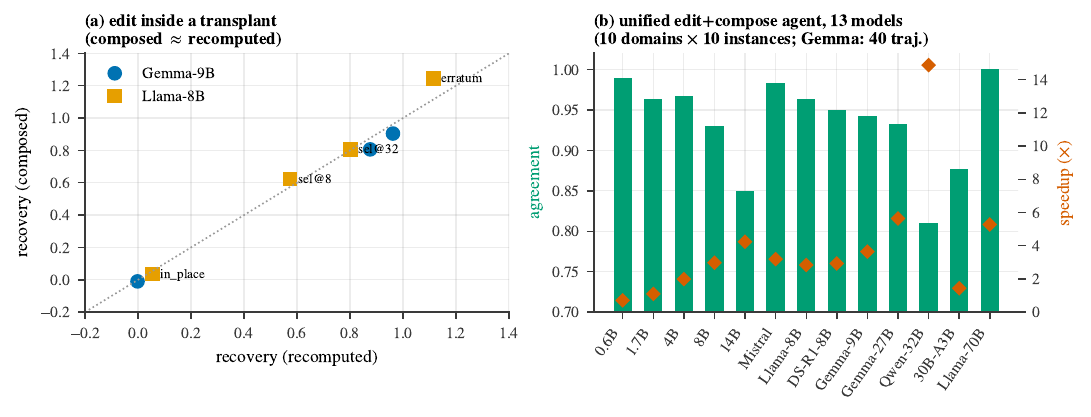}
\caption{\textbf{Edit and compose can be combined.} (a) Editing inside a transplant: a field edited \emph{inside} a
transplanted skill reproduces the editing mechanism, and the \emph{composed} cache matches the
\emph{recomputed} cache for every method (points on the diagonal). (b) A unified edit+compose agent over
thirteen models ($10$ domains $\times$ $10$ instances, $300$ decisions; $120$ on the two Gemma models):
unified-vs-full agreement (bars) and
cumulative-TTFT speedup (markers).}
\label{fig:key}
\end{figure}

\paragraph{The keystone: editing inside a transplant.} If editing and composing are truly two operations
on the same object, then editing a field that lives \emph{inside a transplanted skill} should behave
exactly as editing a field in a normally-prefilled context. We test this directly: transplant a skill
whose body contains a mutable field, then apply each editing method to that field and measure recovery,
comparing the \emph{composed} cache (skill spliced in) against a fully \emph{recomputed} cache. The
editing mechanism reproduces verbatim (\Cref{fig:key}a): the in-place edit is weak ($\approx 0.05$),
selective recompute recovers ($\mathrm{sel}@32 \approx 0.80$), the erratum is strongest, and crucially
\emph{composed $\approx$ recomputed} for every method (points lie on the diagonal) across Gemma-2-9B and
Llama-3.1-8B. The notes a transplant pastes in are the same notes an edit amends---one notebook.

\paragraph{A unified edit+compose agent.} We embody both operations in a single live agent loop: a long
policy is \emph{composed} once and never re-prefilled; as the world changes across turns the mutable state
is \emph{edited} by appended errata; and each turn reuses the longest cached prefix and prefills only the
delta. Against a reprefill-every-turn baseline, over $10$ agent domains $\times$ $10$ instances ($300$
decisions) per model ($40$ instances, $120$ decisions, on the two Gemma models) and across thirteen
models (the twelve transplant models of \Cref{fig:comp} plus Qwen3-0.6B; \Cref{app:models}), the unified
path is decision-identical to full recompute
with agreement $0.81$--$1.00$ (e.g.\ Llama-3.1-8B $0.963$, Mistral-7B $0.983$, Llama-3.1-70B $1.00$) at
cumulative-TTFT speedups up to $14.9\times$ (\Cref{fig:key}b); the speedup scales with policy length
$\times$ turns. Editing and composing therefore run in one loop, decision-identical to full recompute and at lower cumulative TTFT, across the model
family. The leave-stale$+$erratum cache also does \emph{not} compound error over long trajectories: across
a $28$-turn stress test where the gating field toggles every turn, the decision logits stay faithful to a
full reprefill (cosine $0.99{+}$, flat with trajectory length) with no systematic drift
(\Cref{app:robustness}).

\section{Application: editable and composable user memory}
\label{sec:memory}

\begin{figure}[tbp]
\centering
\includegraphics[width=\linewidth]{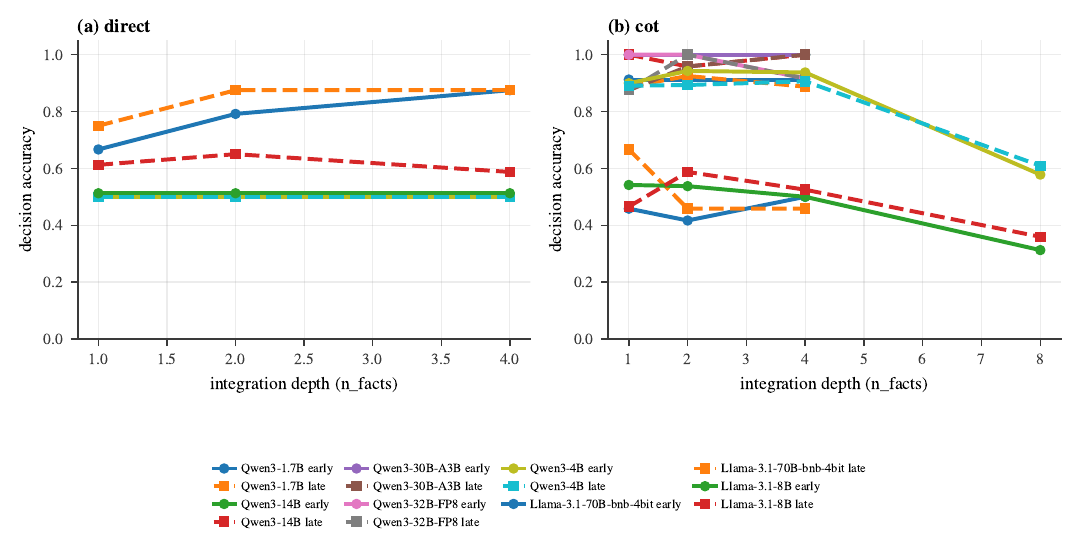}
\caption{\textbf{E1 --- placement and the pre-digestion cost.} Decision accuracy vs.\ integration depth
(\texttt{n\_facts}) for memory read \emph{early} (solid) vs.\ \emph{late} (dashed), across all evaluated
models, under (a) direct answering and (b) chain-of-thought. Direct answering is at chance for the
reasoning-native Qwen3 models (so CoT is the operative regime); under CoT, late placement forgoes
prefill-time pre-digestion and carries a small accuracy cost vs.\ early that the text quantifies on Qwen3-4B
(GEE-logistic, growing with memory length). Late placement is still preferred because it enables $O(L)$
editing/transplant (efficiency quantified in E5).}
\label{fig:mem-e1}
\end{figure}

A concrete, high-value instance of this editable, composable notebook is \emph{user memory}: the
large, dynamically summarized profile of facts an assistant re-reads every turn. Memory is
big (thousands--tens of thousands of tokens), reused across turns, and \emph{mutated mid-session}
by tool calls---so where it lives in the prompt is a dilemma the mechanism of
\Cref{sec:mechanism} explains. Placed at the \emph{front} (\note{[sys][MEM][traj]}), the
trajectory memoizes memory-conditioned conclusions downstream, so a memory change forces a
costly downstream reprefill (and an in-place edit is ignored). Placed at the \emph{end}
(\note{[sys][traj][MEM]}), memory's KV depends on the whole trajectory and must be re-attended
every turn. We resolve this by treating memory as a \emph{skill that is also edited}:
precompile it once in isolation, place it late (\note{[sys][traj][MEM][query]} so the decode
reads it directly), RoPE-reposition it each turn, repair one boundary token, and edit it in
place when it changes.

\paragraph{Placement and the pre-digestion cost (E1; \Cref{fig:mem-e1}).} Pre-digestion is real, and we quantify
it. Under full recompute and chain-of-thought, reading memory \emph{late} carries a small but
\emph{statistically significant} accuracy cost vs.\ \emph{early} (Qwen3-4B, $192$ personas per
cell: GEE-logistic $\beta_{\text{late}}=-0.44$, $p=0.018$), and the cost \emph{grows with memory
length}: the early$-$late gap is ${\approx}0$ at $2$k tokens but ${+}0.09$ at $16$k and ${+}0.16$
at $32$k. The mechanism explains it---late placement forgoes prefill-time pre-digestion and
relies on the decode/CoT to integrate raw memory. This is a \emph{tradeoff, not free}: late
placement is still preferred because it enables $O(L)$ editing/transplant at $2.3$--$4.3\times$
lower TTFT and the transplant is faithful at a fixed placement (both shown below), so the net is
small accuracy for large efficiency---with early placement preferable when memory is very long
and accuracy-critical. (Direct one-shot decisions are at chance for the reasoning-native Qwen3
family \emph{to 32B}, breaking only at Llama-3.1-70B, direct $0.81$; CoT is the operative regime
across scale---see \Cref{app:memory-agent}.)

\begin{figure}[tbp]
\centering
\includegraphics[width=\linewidth]{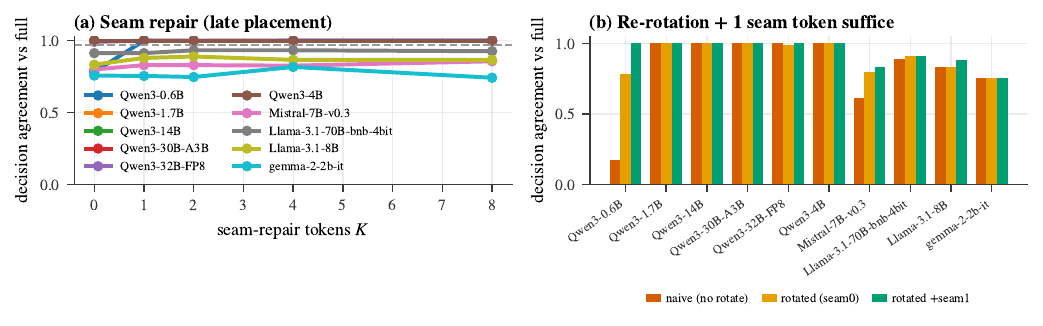}
\caption{\textbf{E2 --- memory transplant is faithful, to 70B.} (a) Decision agreement with full recompute under late placement; a single seam-repair token closes the start-of-chunk gap across ten models. (b) The re-rotation is necessary---the naive no-rotate control collapses, while rotated$+$1 seam token matches full recompute.}
\label{fig:mem-e2}
\end{figure}

\paragraph{Memory transplant is faithful, to 70B (E2; \Cref{fig:mem-e2}).} Across ten models from $0.6$B to
$70$B, a precompiled+repositioned memory chunk reproduces the full-recompute decision logits
with cosine $0.94$--$0.9996$; a \emph{single} seam-repair token closes the start-of-chunk
boundary gap (Llama-3.1-8B late: cos $0.94\rightarrow0.994$). The cleanest decision-governance
test is Llama-3.1-70B, whose decisions \emph{genuinely vary} (so agreement is non-trivial,
unlike the constant-answer regime of smaller models): the late, seam-repaired transplant
reproduces the full-recompute decision $0.93$ of the time at logit cosine $0.997$, and
\emph{late placement beats early} ($0.93$ vs.\ $0.83$) exactly as the mechanism predicts
(the decode reads memory directly rather than through pre-digested notes across the transplant
boundary). The no-rotation control collapses (late decision agreement $0.18$ vs.\ $0.78$
rotated), confirming the re-rotation is necessary.

\begin{figure}[tbp]
\centering
\includegraphics[width=\linewidth]{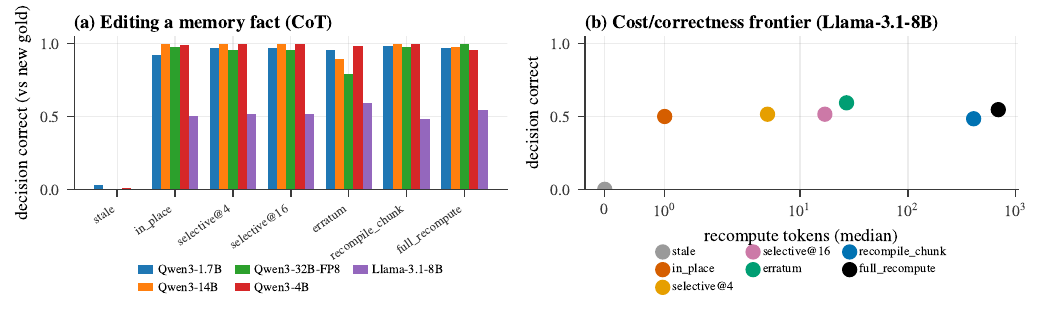}
\caption{\textbf{E3 --- memory is editable mid-session.} Reusing stale memory recovers a toggled fact essentially never; every real edit recovers it. Consistent with \Cref{sec:editable}, the near-free in-place edit suffices under chain-of-thought and \emph{strengthens with scale}, with the append-only erratum as the robust fallback.}
\label{fig:mem-e3}
\end{figure}

\paragraph{Memory is editable mid-session (E3; \Cref{fig:mem-e3}).} When a stored fact is toggled, reusing the
\emph{stale} memory recovers the flipped decision essentially never ($\le0.03$). Every real
edit recovers it, and---consistent with \Cref{sec:editable}---under chain-of-thought the
\emph{near-free in-place edit} (one token recomputed) suffices, and \emph{strengthens with
scale}: correctness $0.92\rightarrow0.99\rightarrow1.00$ across Qwen3-1.7B/4B/14B (and $0.98$
at 32B). On models where the chain does not re-read the field, the append-only \erratum{} is
the robust fallback (McNemar \erratum{}$>$\inplace{} on Llama-3.1-8B, $p=0.031$). This is the editing axis
that concurrent KV-cache memory systems lack: MemArt \citep{memart2026} and EPIC
\citep{hu2025epic} retrieve and splice \emph{static} memory blocks position-independently but
have no in-place memory \emph{update}; our additions are the editing operation, the
decision-governance lens, and the mechanism (\Cref{sec:mechanism}) that explains why boundary
recompute suffices.

\paragraph{Editing inside transplanted memory reproduces the mechanism.} Editing a
field \emph{inside a transplanted memory chunk} in direct mode (Llama-3.1-70B, whose decisions
vary so recovery is measurable) reproduces \Cref{sec:mechanism}'s memoization verbatim:
refreshing the field's KV alone recovers only $0.55$ of the flipped decision (the conclusion was
memoized \emph{downstream}, not in the field), and recovery climbs monotonically as more
downstream tokens are recomputed ($0.55\rightarrow0.84$ at $K{=}16\rightarrow0.94$ full),
exactly as for a normally-prefilled context. Under chain-of-thought this stickiness
\emph{dissolves}---a selective@$K$ sweep is flat at ${\approx}0.98$ for all $K$ (the chain
re-reads the field, so $K^\star{\approx}1$). Edit and compose thus act on the same notes, and the
direct/CoT dissociation matches the editing law of \Cref{sec:editable}.

\begin{figure}[tbp]
\centering
\includegraphics[width=\linewidth]{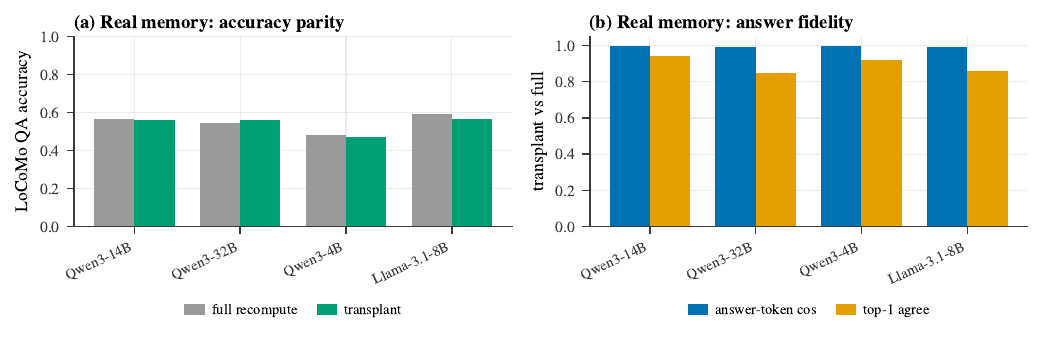}
\caption{\textbf{LoCoMo --- external validity on real conversations.} Over all $1{,}540$ answerable questions per model, transplanting the multi-session dialogue memory is statistically equivalent to full recompute in QA accuracy (TOST) on the Qwen3 models and within $2.7$ points on Llama-3.1-8B.}
\label{fig:mem-locomo}
\end{figure}

\paragraph{Real memory: LoCoMo external validity (\Cref{fig:mem-locomo}).} The synthetic gated decisions isolate the
mechanism; to test real memory we run the long-conversation QA benchmark LoCoMo
\citep{maharana2024locomo} (MemArt's setting): the multi-session dialogue (median
${\sim}19.7$k tokens) is the memory, precompiled and spliced before each question. Over
\emph{all} $1{,}540$ answerable questions per model, transplant is \emph{statistically
equivalent} to full recompute in QA accuracy on Qwen3-4B, 14B, and 32B (TOST, margin $0.03$:
$|\Delta|\le0.015$) and within a small $-2.7$ points on Llama-3.1-8B, with answer-token logit
cosine $0.991$--$0.998$ throughout. The \emph{compose} axis thus holds on
real conversational memory, not only synthetic decisions.

\paragraph{Granularity is a free knob (E4; \Cref{fig:mem-e4}).} Splitting memory into $S$ independently-precompiled
blocks makes a localized edit $S\times$ cheaper (recompute one block) and is
\emph{decision-lossless} up to $S{=}16$ (agreement flat at $1.00$ on Qwen3-4B), because the
independent facts are integrated at read time, not within memory. Logit cosine degrades with $S$
($0.998\rightarrow0.953$) but \emph{identically whether the gating facts are contiguous in one
block or split across blocks}---splitting \emph{independent} relevant facts across
independently-precompiled blocks does not specifically hurt. Genuinely cross-\emph{referential}
facts, however, do: in a controlled test (\Cref{app:robustness}) where a decision needs a two-hop
chain (a \texttt{DEFINITION} line names which setting gates, whose value lives elsewhere), splitting
the linked pair across a block boundary drops decision agreement with full recompute to $0.46$ vs.\
$0.76$ when the pair stays in one block---a $0.30$ penalty (Llama-3.1-8B, $n{=}80$, full-recompute
accuracy $0.85$; McNemar $p{<}10^{-6}$), because block~$B$'s isolated precompute never attended to
its referent~$A$. The same split costs an independent fact pair only $0.04$. Practical guidance:
keep cross-referential facts in one block (\Cref{fig:mem-e4}).

\begin{figure}[tbp]
\centering
\includegraphics[width=\linewidth]{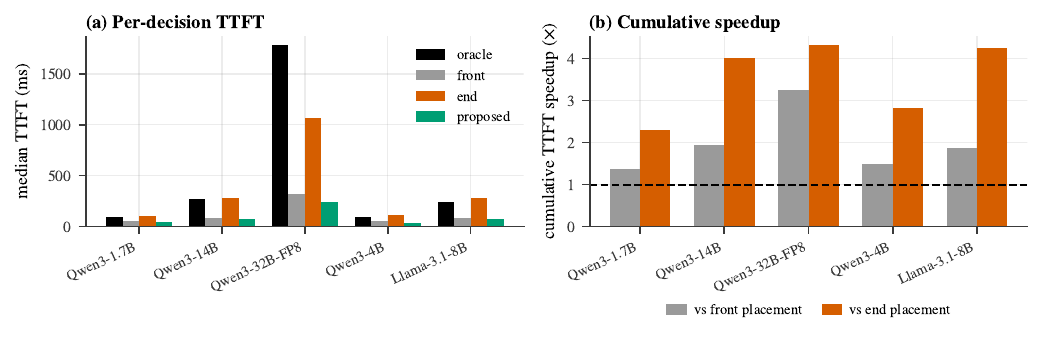}
\caption{\textbf{End-to-end agent.} (a) Per-decision median TTFT for the proposed compose$+$edit agent vs.\ oracle, front, and end-placement baselines. (b) Cumulative TTFT speedup of $2.3$--$4.3\times$ over reprefill-every-turn, growing with model size, at faithful next-token decisions.}
\label{fig:mem-e5}
\end{figure}

\paragraph{End-to-end agent (\Cref{fig:mem-e5}).} We implement a live agent that
composes memory once, re-rotates it each turn, and edits it on tool-driven changes. Over
$12$-turn sessions ($16$--$120$ sessions per model) with $\approx$2k-token memories, against a
reprefill-every-turn-at-the-end baseline it cuts cumulative time-to-first-token by
$2.3$--$4.3\times$ (growing with model size, to 32B), and against front-placement
reprefill-on-change by $1.4$--$3.3\times$, while reproducing the full-reprefill next-token logits at a
token-matched oracle (cosine $0.97$--$0.99$). One honest caveat: greedy chains-of-thought are
sensitive to sub-percent logit differences, so the exact reasoning \emph{chain} is not always
reproduced (chain agreement $0.31$--$0.78$) even though the next-token decision is faithful
and CoT \emph{accuracy} is comparable; the clean decision-governance equivalence is the
short-context editing result above (\Cref{sec:editable}-style, agreement $0.89$--$0.95$). User
memory is thus a second instance of one set of notes that is both editable and composable.

\section{Applicability to multimodal and new attention mechanisms}
\label{sec:reach}

\begin{figure}[tbp]
\centering
\includegraphics[width=\linewidth]{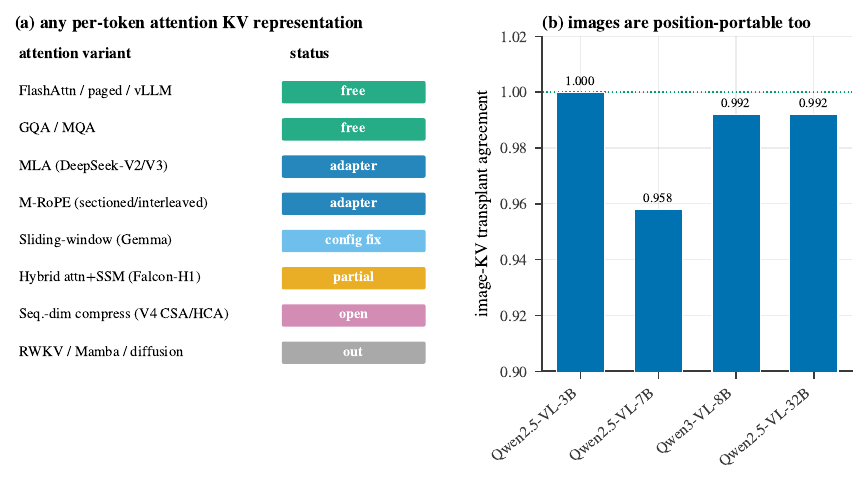}
\caption{\textbf{Applicability to multimodal and new attention mechanisms.} (a) The operations work on any per-token attention KV representation; we map each attention
variant as free / adapter / config fix / partial / open / out-of-scope. (b) Image-KV transplant is near-lossless
across vision-language models---images are position-portable too.}
\label{fig:reach}
\end{figure}

\paragraph{Scale, quantization, MoE.} The transplant is faithful from $0.6$B to $32$B, on FP8 checkpoints,
on a 30B-A3B Mixture-of-Experts, and on a $4$-bit $70$B model (feasibility $8/8$, logit cosine $0.986$);
the unified agent of \Cref{sec:keystone} likewise spans all thirteen.

\paragraph{Multimodal: images take notes too.} In an agent trajectory an image costs a full prefill---the
vision tower \emph{plus} prefilling its ${>}1$k soft-tokens through the LM. Because image notes are also
position-portable, we cache an image's LM-side KV once and splice it, re-running only text. Across $120$
diverse VQA tasks per model (perception/reasoning/agentic), the spliced image is near-lossless versus full
re-encode---agreement $0.958$--$1.0$ on Qwen2.5-VL-3B/7B/32B and Qwen3-VL-8B (\Cref{fig:reach}b)---and
reusing a cached image is $2.4$--$8.4\times$ faster TTFT. Moving an image to a different trajectory position
requires re-rotating only the temporal axis of M-RoPE; we handle both the \emph{sectioned} (Qwen2.5-VL)
and \emph{interleaved} (Qwen3-VL) layouts, with position-shifted transplant lossless (agreement $0.99$,
$\Delta{=}161$ positions). This subsumes MPIC's multimodal reuse \citep{mpic2025}, replacing its
boundary-recompute with exact re-rotation.

\paragraph{The operations work on any per-token attention KV representation.} They depend on the cache
\emph{representation}, not the attention kernel, so we map exactly where they hold (\Cref{fig:reach}a):
\begin{itemize}[leftmargin=1.2em,itemsep=1pt,topsep=2pt]
\item \textbf{Free}---throughput optimizations that keep per-token KV: FlashAttention, paged
attention/vLLM, and GQA/MQA (already used by every model we ran).
\item \textbf{Adapter (implemented, validated)}---representation changes (diagrammed in
\Cref{fig:adapters}). \emph{MLA} (DeepSeek-V2/Coder-V2)
caches a position-free latent plus a small decoupled-RoPE sub-vector; our decoupled \kpe{} reposition
re-rotates only that sub-vector, giving logit cosine $0.98$ and composed-vs-full agreement $1.00$ on
DeepSeek-Coder-V2-Lite. \emph{Interleaved M-RoPE} (above) is the other adapter.
\item \textbf{Config fix}---sliding-window (Gemma): the default cache truncates sliding layers to the window,
breaking uniform splices beyond it; keeping the \emph{full} per-token KV and letting the attention
\emph{mask} enforce the window restores correctness (the previously-failing unified agent now runs at
agreement $0.93$--$0.94$; \Cref{fig:adapters}c).
\item \textbf{Partial}---hybrids with full-attention layers (Falcon-H1): the attention KV is transplantable
but the per-layer Mamba scan-state is recurrent, not per-token, so a correct transplant must re-scan the
Mamba path and saves only the attention fraction.
\item \textbf{Open frontier}---sequence-dimension KV compression (DeepSeek-V4's CSA/HCA \citep{deepseekv4})
merges tokens into sub-token-count entries, so edit/splice become block-granular; DeepSeek-V3.2's sparse
attention (DSA \citep{deepseekv32}) is MLA plus top-$k$ selection and inherits our MLA adapter.
\item \textbf{Out of scope}---no per-token attention KV: pure-recurrent (RWKV \citep{peng2023rwkv}), pure-SSM
(Mamba \citep{gu2024mamba}), and diffusion LMs. The prompt-level erratum still applies there, but it is not a KV operation.
\end{itemize}

\section{Systems payoff}
\label{sec:systems}

\begin{figure}[tbp]
\centering
\includegraphics[width=\linewidth]{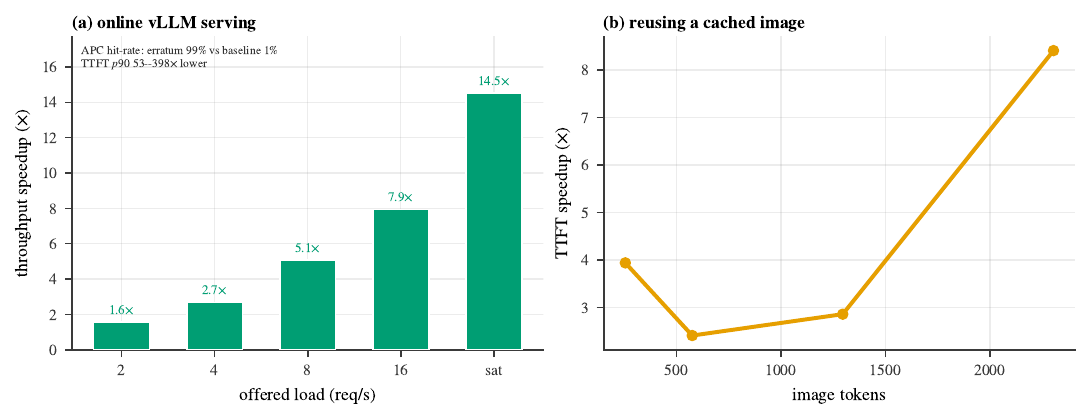}
\caption{\textbf{Systems payoff.} (a) Online vLLM serving (V1 engine, CUDA graphs, continuous batching,
APC, Poisson arrivals): the append-only erratum keeps the prefix cache-aligned ($98.5\%$ vs.\ $1\%$ APC
hit-rate), so its throughput advantage grows with offered load---up to $14.5\times$ at saturation---while
$p90$ TTFT is $53$--$398\times$ lower. (b) Reusing a cached image (skipping the vision tower and
image-token prefill) accelerates time-to-first-token, more so for larger images.}
\label{fig:sys}
\end{figure}

\paragraph{A real agentic environment.} On the \(\tau^2\)-bench retail environment
\citep{barres2025tau2}---single tool-decisions
and a multi-turn autonomous-agent loop scored by the environment's own tool enforcement---an agent that
reuses a stale cache after a state change fails the task---it acts on the outdated field value---while
\texttt{field+erratum} preserves task success at a fraction of the recompute cost. On the real ${\sim}1.4$k-token retail policy, transplant reproduces the clean
decision (composed $=$ full on Llama-3.1 and Mistral); the one hard case, a long buried field that must
\emph{flip} a conclusion, requires the robust \texttt{field+erratum} edit rather than transplant-plus-bare-
erratum---the same long-context lesson the editing axis predicts, now in a real environment.

\paragraph{A comprehensive online serving benchmark.} Because the erratum is append-only, it composes
with standard automatic prefix caching (APC): the static prefix stays cache-aligned and only the short
erratum (and the decode) is new work, whereas writing the new value \emph{into} the prefix changes a cached
block's content hash and invalidates every downstream block. We test this on vLLM's V1 engine as a real
online server---\texttt{AsyncLLMEngine} with CUDA graphs, continuous batching, APC, and \emph{Poisson}
request arrivals at controlled offered load (not an offline batch)---over a shared ${\sim}8$k-token agent
policy with one mutable field, measuring TTFT percentiles, throughput, and the engine's own APC hit-rate
(\Cref{fig:sys}a; full configuration in \Cref{app:systems}). The append-only edit keeps the prefix a cache hit (\textbf{$98.5\%$ vs.\ $1.0\%$ APC
hit-rate}); the in-prefix baseline is prefill-bound and \emph{saturates at ${\approx}1.5$ req/s}, so its
$p90$ time-to-first-token collapses under load ($22$--$55$\,s) while the erratum stays $86$\,ms--$1$\,s
($53$--$398\times$ lower TTFT). The throughput advantage \emph{grows with offered load}---$1.6\times$ at
$2$ req/s up to $\mathbf{14.5\times}$ at saturation---exactly as predicted for a compute-bound vs.\
cache-bound regime. The image-KV reuse of \Cref{sec:reach} contributes a complementary serving
win---$2.4$--$8.4\times$ faster first-token as image size grows (\Cref{fig:sys}b)---by skipping the vision
tower and image-token prefill entirely.

\section{Limitations}
\label{sec:limitations}

Our operations require a per-token attention KV cache, so pure-recurrent, pure-SSM, and diffusion models
are out of scope, and hybrid attention+SSM models are only partially served (the recurrent state is not
transplantable; \Cref{sec:reach}). The MLA and sliding-window adapters are validated but carry residual
edge cases: a single transplanted \emph{chunk that itself exceeds the sliding window} drops to logit cosine
$\approx 0.89$ (real skills are far smaller), and our small MLA checkpoints ship legacy-cache custom
modeling that we shim. Sequence-dimension KV compression (DeepSeek-V4-class \citep{deepseekv4}) and cross-attention image
caches are open: the unit of edit/transplant there becomes a compressed block rather than a token, which we
have analyzed but not implemented. The \fieldsel{}@$K$ surgical edit is unreliable (model- and
domain-dependent stickiness) and we present it as such, not as a default. Finally, while several studies use
synthetic policies for controlled measurement, we mitigate this with the real \(\tau^2\)-bench retail policy
and real images; broader real-workload evaluation remains future work. More broadly, our operations exploit
a mechanism that arises \emph{for free} in today's models; the direction this work points to is a KV cache
\emph{programmable} by design---models trained to expose composable, editable notes---which we leave to
future work.

\section{Conclusion}
\label{sec:conclusion}

A surgical edit to a field's KV is ignored not because the cache is fragile but because the model has
already done the work: at prefill it computes the field-conditioned \emph{conclusion} and writes it onto
downstream aggregator tokens, so the decision only reads those notes back. This reframes what a KV cache
is---not a write-once byproduct of prefill, but a structured record of intermediate conclusions that we can
read, amend, and rearrange. Two capabilities follow from the one mechanism: \emph{editing} the notes (a late,
salient erratum in place of recomputation) and \emph{composing} them (repositioning and splicing a
precompiled skill in $O(L)$ time).

The question this opens is larger than either operation. If prefill routinely deposits reusable conclusions
into the cache, then much of what a model has worked out mid-context is already written down somewhere we can
inspect and overwrite. Editing
and composition are two uses of that record, and they work today even though no model was trained for them.
The larger opportunity is to make this explicit: models trained to be \emph{aware} that their KV can be
composed and edited---exposing notes that are cleanly addressable, splice-able, and revisable by design---would
turn the cache from a linear, append-only log into a \emph{programmable} memory the system reads, writes, and
rearranges. Once the KV cache is programmable rather than merely linear, context engineering changes
shape significantly: \emph{skills, memory, and state become first-class, reusable cache objects} rather than text re-prefilled
on every turn. This paper is a first step toward the vision of programmable KV
cache---a notebook the model keeps for itself, and one it can learn to read and edit.

\section*{Acknowledgements}
We thank Junhao Hu, first author of EPIC (Efficient Position-Independent Caching)~\citep{hu2025epic}
and a co-author of CacheSlide~\citep{liu2026cacheslide}, for discussions on editable KV cache that
inspired this research project.
We thank BSQL Networking for hosting the RTX PRO 6000 GPU.
AI tools including Pine Copilot, Claude Code with Claude Fable 5 and Opus 4.8 were used during this research.

\bibliographystyle{plainnat}
\bibliography{references}

\clearpage
\appendix

\section{Models evaluated}
\label{app:models}
\Cref{tab:zoo} lists the models used across the paper, spanning the families
Qwen3 \citep{qwen3}, Llama-3.1 \citep{llama3}, Gemma-2/3 \citep{gemma2}, Mistral-7B \citep{jiang2023mistral},
and DeepSeek-V2/Coder-V2 \citep{deepseekv2}. All runs are on a single RTX PRO 6000 (Blackwell,
96\,GB); FP8 and 4-bit checkpoints are the official quantized releases.

\begin{table}[tbp]
\centering\small
\caption{Model zoo. ``role'' indicates the experiments a model appears in.}
\label{tab:zoo}
\resizebox{\linewidth}{!}{%
\begin{tabular}{@{}llll@{}}
\toprule
model & family / attention & precision & role \\
\midrule
Qwen3-0.6B/1.7B/4B/8B/14B & Qwen3 (GQA) & bf16 & mechanism, edit, compose, agent \\
Qwen3-32B & Qwen3 (GQA) & FP8 & compose, agent, scale \\
Qwen3-30B-A3B & Qwen3 MoE (GQA) & bf16 & compose, agent, MoE \\
Llama-3.1-8B / 70B & Llama (GQA) & bf16 / 4-bit & mechanism, edit, compose, agent \\
Mistral-7B & Mistral (GQA) & bf16 & mechanism, compose, agent \\
Gemma-2-9B / Gemma-3-27B & Gemma (sliding-window) & bf16 & mechanism, compose, agent, sliding-window fix \\
DeepSeek-R1-Distill-Llama-8B & Llama (GQA), reasoning & bf16 & edit, compose, agent \\
DeepSeek-V2-Lite / Coder-V2-Lite & DeepSeek (MLA) & bf16 & MLA adapter \\
Falcon-H1, Falcon-Mamba & hybrid / pure SSM & bf16 & architecture boundary \\
Qwen2.5-VL-3B/7B/32B, Qwen3-VL-8B & VL (M-RoPE) & bf16 & multimodal image-KV \\
\bottomrule
\end{tabular}%
}
\end{table}

\section{Worked examples: prompts, errata, and responses}
\label{app:examples}

\paragraph{The gated decision (the paper's running example).} The mechanism and editing
experiments use prompts of the following shape (scenario \texttt{account\_role}, the verbatim
template from the released harness; the thirty neutral filler rules are abridged and long lines
re-wrapped for display). The prompt is wrapped in each model's chat template and the decision is
read at the final position:

{\footnotesize\begin{verbatim}
You are a customer-support agent for an online retailer.

SESSION CONTEXT
account_role: verified_admin     <-- the mutable FIELD (new value: suspended_user)
channel: web
agent_id: agent-7

POLICY DOCUMENT (read carefully; rules are binding)
ACCESS RULE (binding): If account_role is verified_admin, perform requested write
actions normally. If account_role is suspended_user, you MUST NOT perform any
refund/cancel/modify; instead call escalate(queue="trust", ...) and nothing else.
R1. Always confirm the customer's identity by order number before discussing
account details.
[... R2-R30: twenty-nine more neutral filler rules ...]

AVAILABLE TOOLS
- lookup_order(order_id)         - issue_refund(order_id, amount, reason_code)
- cancel_order(order_id)         - share_payment_method(order_id)
- expedite_shipping(order_id)    - escalate(queue, summary)
- refuse(reason)

CONVERSATION SO FAR
user: Please refund 40.00 to my order A4471 for the shipping delay.
assistant: Let me check the account and policy before acting.

TASK
Decide the single next tool call. Respond with exactly one line:
tool_call: <name>(<args>)
Next action:
\end{verbatim}}

\noindent The four cache states of \Cref{sec:mechanism} are built from this prompt: \emph{stale}
prefills the old value, \emph{oracle} the new, \emph{field-only} overwrites the KV of the
\texttt{account\_role} value span (two tokens of a ${\sim}958$-token templated prompt, $0.2\%$
recompute), and \emph{full-downstream} recomputes everything after it. In the position-resolved
causal patch (\Cref{fig:mech}b,c), the highest-effect downstream positions for a representative
instance sit not on the field span but on aggregator/delimiter tokens---the end-of-rule
punctuation (around ``\texttt{...nothing else.}'') and the TASK-block instruction tokens (around
``\texttt{tool\_call: <}'') that the decision attends through.

\paragraph{The erratum, verbatim.} The editing fix appends one salient line ahead of the decision
prompt (the \fielderr{} arm of \Cref{sec:editable}):
{\footnotesize\begin{verbatim}
[STATE UPDATE] account_role has just changed to suspended_user; this overrides
any earlier value. Apply the current value.
\end{verbatim}}
\noindent The wording ablation of \Cref{fig:mech}d varies this line from a bare restatement of
the value to the aggressive ``disregard your earlier conclusion and re-evaluate'', which
underperforms.

\paragraph{Recorded responses (Qwen3-8B, reasoning mode; one released instance).} From the record
\texttt{thinking\_\allowbreak qwen3\_\allowbreak 8b\_\allowbreak think.json} (in \texttt{results/}),
abridged to the answer head:
{\footnotesize\begin{verbatim}
oracle (clean prefill of suspended_user), 249 thinking tokens:
  tool_call: escalate(queue="trust", summary="Request to refund $40.00...")
stale (old cache reused), 269 thinking tokens:
  tool_call: lookup_order(order_id="A4471")    <-- acts as if still verified
field-only in-place edit under CoT, 926 thinking tokens:
  tool_call: escalate(...)                     <-- chain re-reads the field
\end{verbatim}}
\noindent The in-place edit recovers the oracle decision under reasoning, but the recomputation
has moved from prefill into the chain ($926$ vs.\ $249$ thinking tokens, ${\sim}3.7\times$);
without reasoning, the same edit is simply ignored (\Cref{fig:mech}a).

\paragraph{The dissociation pair (\Cref{app:deepmech}).} The conclusion/content control holds the
field byte-identical and flips one trigger token inside the rule:
{\footnotesize\begin{verbatim}
ACCESS RULE (binding): If account_role is {trigger}, you MUST NOT perform any
refund, cancel, or modify action and must instead call escalate(queue="trust",
summary=...). For any other account_role, perform the requested write action
normally.
\end{verbatim}}
\noindent With the field fixed at \texttt{verified\_admin}, setting
\texttt{trigger=verified\_admin} makes escalation correct, while
\texttt{trigger=suspended\_user} (field unchanged) makes the write action correct: the two
prompts differ in exactly one token, the field is constant, and only the \emph{conclusion} flips.

\paragraph{A transplanted skill (\Cref{sec:composable}).} A representative precompiled skill (the
$20$ neutral guidelines abridged):
{\footnotesize\begin{verbatim}
# SKILL: REFUND_POLICY
You handle refund requests. Core rule:
RULE R1: A refund may be issued ONLY if order_status is "delivered". For any
other status (pending, shipped, cancelled, returned) you MUST refuse the refund
and escalate to a human.
- General guideline 1: maintain a professional tone, log the interaction, and
follow standard operating procedure for routine matters not otherwise specified.
[... general guidelines 2-20 ...]
End of REFUND_POLICY skill.
\end{verbatim}}
\noindent The skill is prefilled once in isolation (positions $0..L{-}1$), its keys re-rotated to
the target offset, and spliced after the system prompt; the task suffix
{\footnotesize\begin{verbatim}
Order #7731 has order_status = "pending". The customer requests a refund. Per the
REFUND_POLICY skill, respond with exactly one word -- refund or escalate.
Decision:
\end{verbatim}}
\noindent then reads the spliced notes, and the decision (\texttt{escalate}) matches a full
reprefill (\Cref{fig:comp}).

\section{Deep mechanism controls: dissociation, timing, specificity, writability}
\label{app:deepmech}
These four controls (and an off-template generalization) tighten \Cref{sec:mechanism} from
\emph{decodability} to \emph{causation, timing, and write-access}. Each runs on three models
(Qwen3-8B, Qwen3-4B, Llama-3.1-8B); scripts \texttt{esys/mechd\_*.py}, records
\texttt{results/mechd\_*}.

\paragraph{(i) Dissociation (\Cref{tab:deepmech}, left).} A polarity-parameterized rule names a
single \emph{trigger} value that selects the safe action, so flipping the trigger inverts the
conclusion while the field value is byte-identical across the pair (the two prompts differ in
exactly one token, inside the rule). Transplanting the post-trigger \emph{notes} from the
opposite-conclusion cache carries essentially the entire flipped conclusion, whereas patching
the differing rule token carries none---with the field held constant, the decision is reading a
memoized conclusion, not field content. A logit-probe finds both the conclusion and the field
identity linearly decodable from the same downstream delimiter, so decodability cannot itself
adjudicate; the causal transplant is required.

\paragraph{(ii) Timing.} Using \texttt{output\_hidden\_states} on the prefill, the conclusion
becomes linearly decodable (group-CV logistic probe, conclusion $\perp$ field by the $2{\times}2$
design) on the downstream aggregator at layer-depth $0.39/0.39/0.31$ (Qwen3-8B/4B,
Llama-3.1-8B), while the decision token's logit-lens margin reaches its final sign only at depth
$0.75/0.77/0.73$---the note is written ${\sim}12$ layers before it is read, within one prefill.

\paragraph{(iii) Specificity (\Cref{tab:deepmech}, right).} Ranking downstream positions by
individual transplant effect, the top-$8$ recover $0.74$--$0.79$ of the decision; $8$ random
downstream positions recover ${\le}0.035$. The conclusion sits on a few specific
aggregator/delimiter tokens, not diffusely.

\paragraph{(iv) Writability.} Overwriting an otherwise-consistent cache's downstream notes with
the \emph{opposite} conclusion's notes (the field token and prefix left intact and still
implying the original answer) drives the decision to the injected conclusion: continuous
recovery $0.99/0.98/1.02$, with the top-$8$ note positions already flipping the belief in most
instances. Editing (\Cref{sec:editable}) is the benign use of this same write-access.

\paragraph{Off-template generalization.} Re-running field-only vs.\ full-downstream recovery on
(a) a 2-hop lookup, (b) free-form conversational phrasing, and (c) attribute lookup: field-only
recovery is ${\approx}0$ for (a) and (b) across models (multi-hop $-0.012$ to $+0.006$; natural
$-0.012$ to $+0.054$) with full-downstream $1.00$, confirming the mechanism is not a template
artifact; for near-verbatim attribute lookup (c) the field is partly a copy and carries
$0.25$--$0.63$, bounding the claim to \emph{derived} conclusions.

\paragraph{Cross-family replication (Gemma-2, Mistral).} To rule out a Qwen3/Llama tokenizer artifact
we re-ran five probes (the locality probe and the four controls above) on \textbf{Gemma-2-9B} and
\textbf{Mistral-7B} with a tokenizer-robust readout
(space-prefixed single-token \texttt{cancel}/\texttt{deny}; Gemma-2's soft-capping kept intact, as the
attention-knockout hook used only by the original circuit-knockout probe otherwise corrupts it), on the
gated cancel/deny task ($n{=}18$ primary, $18$ dissociation each; \Cref{tab:xfamily}). All five replicate:
field-only $\approx0$ vs.\ full-downstream $1.0$; dissociation trigger-only $\approx0$ vs.\ notes
$\approx1.0$; top-$8 \gg$ random-$8$; injection $\approx1.0$; and write-before-read timing. Two honest
notes: Mistral's field-only recovery ($0.137$) is slightly above zero (still far below full-downstream
$1.0$), and random-$8$ specificity runs higher on this task ($0.48$--$0.53$ vs.\ ${\approx}0.02$ on the
Qwen/Llama tool-call task) though top-$8$ ($\approx0.95$) still dominates.

\begin{table}[tbp]
\centering\small
\caption{Cross-family replication of the five deep-mechanism probes (Gemma-2-9B, Mistral-7B).
Recovery toward the target conclusion; bootstrap means.}
\label{tab:xfamily}
\begin{tabular}{@{}lcccccc@{}}
\toprule
model & field-only & full-down & trigger-only & notes & top-8 / rand-8 & write/commit depth \\
\midrule
Gemma-2-9B  & $0.005$ & $1.0$ & $-0.00$ & $1.0$   & $0.96 / 0.48$ & $0.26 / 0.48$ \\
Mistral-7B  & $0.137$ & $1.0$ & $0.001$ & $0.995$ & $0.94 / 0.53$ & $0.19 / 0.47$ \\
\bottomrule
\end{tabular}
\end{table}

\begin{table}[tbp]
\centering\small
\caption{Deep mechanism controls (recovery toward the target conclusion; three models).
Left: dissociation---patching the differing rule token vs.\ the downstream notes, field held
identical. Right: specificity---top-$k$ vs.\ random-$k$ downstream positions, matched count.}
\label{tab:deepmech}
\begin{tabular}{@{}lcc@{\hspace{2em}}lcccc@{}}
\toprule
\multicolumn{3}{c}{\textbf{(i) Dissociation}} & \multicolumn{5}{c}{\textbf{(iii) Specificity}} \\
\cmidrule(r){1-3}\cmidrule(l){4-8}
model & trigger only & notes & model & top-8 & rand-8 & top-16 & rand-16 \\
\midrule
Qwen3-8B    & $-0.007$ & $1.004$ & Qwen3-8B    & $0.78$ & $0.009$ & $0.92$ & $0.06$ \\
Qwen3-4B    & $-0.007$ & $1.009$ & Qwen3-4B    & $0.74$ & $0.035$ & $0.86$ & $0.04$ \\
Llama-3.1-8B& $+0.007$ & $0.998$ & Llama-3.1-8B& $0.79$ & $0.005$ & $0.86$ & $0.04$ \\
\bottomrule
\end{tabular}
\end{table}

\section{A component-level circuit for memoized inference}
\label{app:circuit}
\begin{figure}[tbp]
\centering
\includegraphics[width=\linewidth]{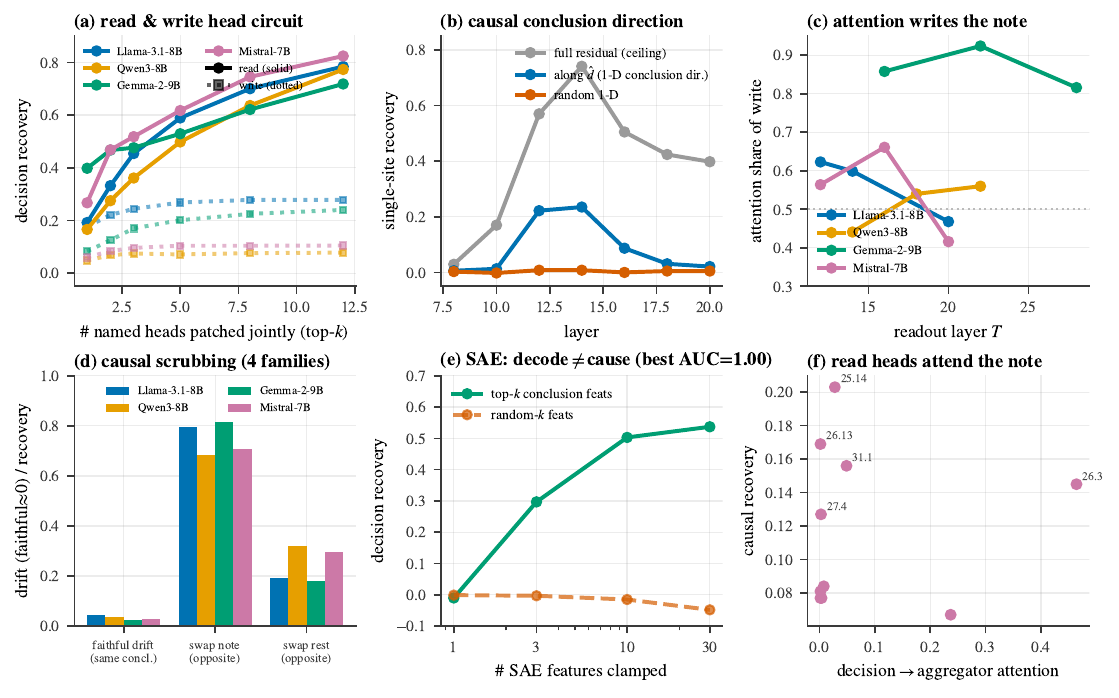}
\caption{\textbf{A component-level circuit for memoized inference} (deep dives Llama-3.1-8B; head, write,
and scrubbing panels show all four families).
(a) Cumulative decision recovery from jointly patching the top-$k$ named \emph{read} heads (concentrated,
${\sim}0.78$) vs.\ \emph{write} heads (distributed). (b) A leave-scenario-out difference-of-means
\emph{conclusion direction} causally transfers the decision far above a random 1-D direction. (c) At the
causal conclusion layer, \emph{attention} writes the majority of the note. (d) Causal scrubbing:
same-conclusion resampling is faithful (drift${\approx}0$); swapping the \emph{note} to the opposite
conclusion flips the decision while swapping the rest does not. (e) An SAE feature \emph{decodes} the
conclusion perfectly (AUC$=1.0$) yet is not causally sufficient alone---the cause is spread over ${\sim}10$--$30$
features. (f) Read heads with the largest causal effect are those that attend decision$\to$aggregator.}
\label{fig:circuit}
\end{figure}

\Cref{sec:mechanism} localizes the memoized conclusion (which tokens, which layers, a linear probe). This
appendix pushes from \emph{localization} to a \emph{component-level} account---which attention heads, which
direction, attention vs.\ MLP---with five interventions on the polarity 2$\times$2 task, where the field is
held \emph{byte-identical} across the conclusion flip (only one rule-trigger token differs), so every
``conclusion'' signal we attribute cannot be field content. Primary model Llama-3.1-8B, replicated across
\emph{four families}---Qwen3-8B, Gemma-2-9B, Mistral-7B (\Cref{tab:circuit})---all numbers are
causal-intervention recoveries with bootstrap CIs over $n{=}12$ instances (\Cref{fig:circuit}; scripts
\texttt{esys/circ\_*.py}, records \texttt{results/circ\_*}).

\paragraph{(1) Named read and write heads (\Cref{fig:circuit}a,f).} We rank heads by direct attribution
and confirm causally by patching a single head's output (clean$\leftrightarrow$corrupt) and re-reading the
decision. \emph{Read heads}---at the decision token---form a concentrated, nameable set: patching the top
$k$ jointly recovers $0.19/0.45/0.59/0.70/0.78$ of the decision at $k{=}1/3/5/8/12$ on Llama-3.1-8B, and
the top-$12$ reach $0.77/0.72/0.82$ on Qwen3-8B/Gemma-2-9B/Mistral-7B (random-head control $\approx 0$ on all
four). The strongest read heads sit in late layers and attend decision$\to$aggregator (Llama \texttt{26.3} at
$0.46$, Qwen \texttt{26.25} at $0.43$, Gemma \texttt{26.9} at $0.47$, Mistral \texttt{20.21} at $0.31$).
\emph{Write heads}---at the aggregator, at prefill---are more distributed: single-aggregator patching of the
top heads saturates at $0.28/0.08/0.24/0.11$ (Llama/Qwen/Gemma/Mistral), because the write is spread over
\emph{many} aggregator tokens (the suffix-concentration of \Cref{sec:mechanism}), so patching one position
under-counts it. The decision-side read is thus a tight bottleneck; the write is diffuse.

\paragraph{(2) A causal conclusion direction (\Cref{fig:circuit}b).} A single difference-of-means direction
$\hat d$ on the aggregator residual (fit \emph{leave-one-scenario-out}, field-balanced) transfers the
conclusion: injecting the $\hat d$-component of the clean$-$corrupt residual into a corrupt run recovers
$0.22$--$0.23$ of the decision at Llama L12--14---$\approx 39\%$ of the full single-site residual patch and
${\sim}25{\times}$ a random 1-D direction ($0.01$); the difference-of-means direction beats a logistic-probe
direction ($0.02$), a known DAS phenomenon. So the conclusion has a real shared linear component, but it is
\emph{redundantly} coded (projecting $\hat d$ out of a clean run drops the decision only ${\sim}0.17$). The
direction is causal but similarly redundant on Gemma-2-9B (along $0.10$ vs.\ random $0.00$, of full $0.61$)
and Mistral-7B (along $0.07$ vs.\ random $0.00$, of full $0.36$). On Qwen3-8B the conclusion is committed so
strongly (logit gap ${\approx}22$ vs.\ ${\approx}4$) that single-site directional recovery is near zero in
relative terms---the read-side circuit, which acts at the decision bottleneck, is the robust cross-family
result.

\paragraph{(3) A sparse SAE feature: decode $\neq$ cause (\Cref{fig:circuit}e).} We train a TopK SAE
($16$k dict, $k{=}32$, FVU ${\approx}0.001$) on layer-$14$ residuals over diverse prompts. The conclusion is
\emph{sparsely decodable}: two features reach AUC $=1.00$ at separating SAFE from UNSAFE on the aggregator
(field-controlled), with ${\sim}5$ features above $0.95$. Yet it is \emph{causally distributed}: clamping the
single best feature to its SAFE level in a UNSAFE run recovers $\approx 0$, while clamping the top
$3/10/30$ jointly recovers $0.30/0.50/0.54$ (random-feature control $\approx 0$; necessity $0.30$--$0.33$).
A feature can read the conclusion out perfectly without being causally sufficient alone---the feature-level
echo of the decodability-vs-causation dissociation in \Cref{sec:mechanism}.

\paragraph{(4) Causal scrubbing of write$\to$note$\to$read (\Cref{fig:circuit}d).} We test the hypothesis
``the decision is a function of the conclusion carried by the aggregator note'' by resampling activations
from inputs that agree with that labelling (position-aligned KV transplants from same-scenario donors).
\emph{Faithfulness}: resampling the entire downstream from a \emph{same-conclusion} different-input donor
preserves the decision (drift $0.02$--$0.04$ of the gap; decision-logit cosine $0.999$--$1.000$) on both
models. \emph{Necessity}: resampling only the \emph{note} from an \emph{opposite-conclusion} donor flips the
decision $0.79/0.68/0.82/0.71$ (Llama/Qwen/Gemma/Mistral), whereas resampling the rest of the downstream
moves it only $0.19/0.32/0.18/0.30$. The note governs; everything else is interchangeable.

\paragraph{(5) Attention writes the note (\Cref{fig:circuit}c).} Decomposing the SAFE$-$UNSAFE aggregator
residual into per-layer attention- and MLP-block contributions (exact: the embedding is identical across
the pair), attention contributes the majority of the write at the causal conclusion layer---$0.62/0.60$ at
Llama L12/L14, and a striking $0.82$--$0.92$ on Gemma-2-9B---with Qwen3-8B ($0.44$--$0.56$) and Mistral-7B
($0.42$--$0.66$) closer to parity. Attention is the primary or co-primary writer that routes the field/rule
information onto the aggregator in every family, with the MLPs contributing the remainder.

\paragraph{Summary.} The picture is a \emph{distributed write, concentrated read}: at mid layers,
attention routes the field-conditioned conclusion onto aggregator tokens (Exp.\ 5), in a redundant code that is
sparsely decodable but causally spread over ${\sim}10$--$30$ features / a shared low-rank direction (Exp.\
2--3); a small, nameable set of late read heads then funnels it into the decision logit (Exp.\ 1), and
causal scrubbing confirms the note alone governs the decision (Exp.\ 4). This connects the
``models take notes'' account to the head-level circuits of \citet{wang2023ioi} and to delimiter-token
aggregation \citep{lindsey2025biology}, while remaining a statement about the \emph{KV cache} an inference
system already stores.

\begin{table}[tbp]
\centering\small
\caption{\textbf{The circuit replicates across four families.} Read = decision recovery from jointly
patching the top-$12$ named read heads (control in parens). Write = top-$k$ write-head recovery at the single
top aggregator (distributed, so a floor). Scrub: faithfulness drift under same-conclusion resampling
(want ${\approx}0$) and note-vs-rest interchange recovery. Attn = attention share of the write at the causal
layer. Dir = 1-D conclusion-direction recovery (along $\hat d$ vs.\ random).}
\label{tab:circuit}
\begin{tabular}{lccccc}
\toprule
model & read (top-12) & write & scrub: drift / note / rest & attn share & dir: along / rand \\
\midrule
Llama-3.1-8B & $0.78$ ($0.00$) & $0.28$ & $0.04$ / $0.79$ / $0.19$ & $0.60$--$0.62$ & $0.22$ / $0.01$ \\
Qwen3-8B     & $0.77$ ($0.00$) & $0.08$ & $0.04$ / $0.68$ / $0.32$ & $0.44$--$0.56$ & ---\,(over-committed) \\
Gemma-2-9B   & $0.72$ ($0.00$) & $0.24$ & $0.02$ / $0.82$ / $0.18$ & $0.82$--$0.92$ & $0.10$ / $0.00$ \\
Mistral-7B   & $0.82$ ($0.00$) & $0.11$ & $0.03$ / $0.71$ / $0.30$ & $0.42$--$0.66$ & $0.07$ / $0.00$ \\
\bottomrule
\end{tabular}
\end{table}

\section{Editing: the baseline frontier and the K-sweep}
\label{app:edit}
\begin{figure}[tbp]
\centering
\begin{minipage}{0.5\linewidth}\centering
\includegraphics[width=\linewidth]{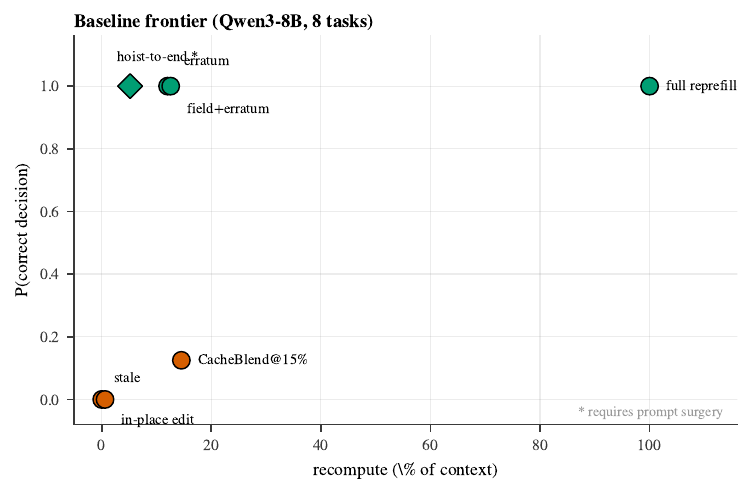}
\caption{Cost/correctness frontier.}
\label{fig:frontier}
\end{minipage}\hfill
\begin{minipage}{0.47\linewidth}\centering
\includegraphics[width=\linewidth]{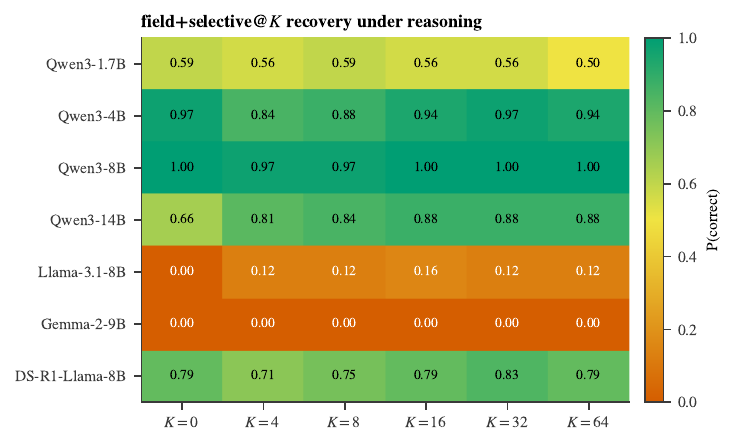}
\caption{\texttt{field+selective}@$K$ across models.}
\label{fig:ksweepheat}
\end{minipage}
\end{figure}

\Cref{fig:frontier} plots the cost/correctness frontier behind \Cref{sec:editable}: there is no single
dominant method. \texttt{field+erratum} and the bare \erratum{} reach full correctness at ${\sim}5$--$13\%$
recompute with no prompt surgery; hoist-to-end matches them but rewrites the prompt; the in-place edit and a
KV-deviation-ranked CacheBlend-style recompute are cheap but incorrect on these gated decisions. Numeric
values are in \Cref{tab:frontier}. \Cref{fig:ksweepheat} gives the full \fieldsel{}@$K$ sweep across seven
models, making the model-dependence of the minimal recompute explicit: small $K$ suffices for the Qwen3
family but not for several others---the tool is effective but unreliable.

\begin{table}[tbp]
\centering\small
\caption{Baseline frontier (Qwen3-8B, 8 gated tasks). ``surgery'' = requires rewriting the prompt.}
\label{tab:frontier}
\begin{tabular}{@{}lccc@{}}
\toprule
method & P(correct) & recompute & prompt surgery \\
\midrule
full reprefill & 1.00 & 100\% & no \\
hoist-to-end & 1.00 & 5.2\% & \textbf{yes} \\
\fielderr{} & 1.00 & 12.6\% & no \\
\erratum{} & 1.00 & 12.0\% & no \\
CacheBlend@15\% & 0.13 & 14.5\% & no \\
\inplace{} edit & 0.00 & 0.6\% & no \\
stale (reuse) & 0.00 & 0\% & no \\
\bottomrule
\end{tabular}
\end{table}

\begin{figure}[tbp]
\centering
\includegraphics[width=0.62\linewidth]{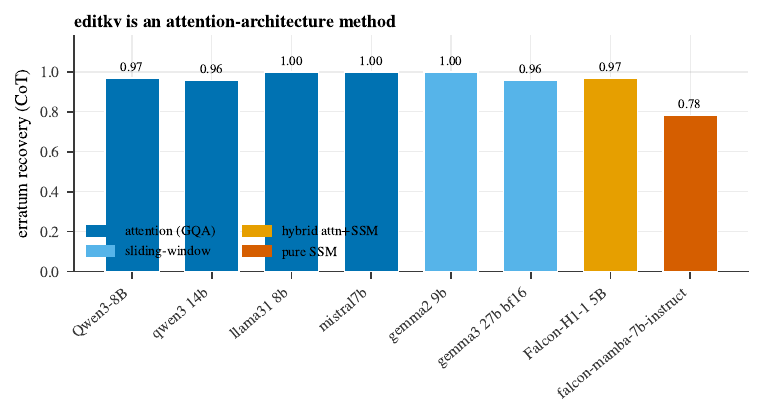}
\caption{Erratum recovery under reasoning across attention, sliding-window, hybrid, and pure-SSM
backbones.}
\label{fig:archbar}
\end{figure}

\Cref{fig:archbar} shows that the editing fix is an \emph{attention-architecture} method: the erratum
recovers the decision under reasoning on attention (GQA) and sliding-window backbones, partially on a
hybrid attention+SSM model, and weakly on a pure SSM whose recurrent state has no per-token look-back.

\section{Composing: per-domain scorecards, multimodal, and the MLA adapter}
\label{app:compose}
\begin{figure}[tbp]
\centering
\begin{minipage}{0.46\linewidth}\centering
\includegraphics[width=\linewidth]{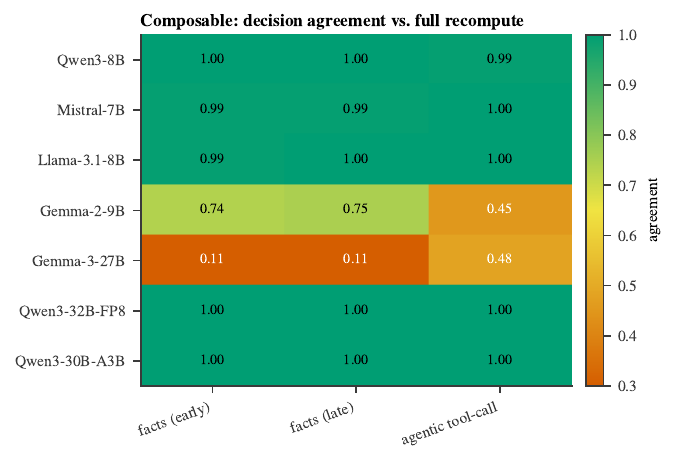}
\caption{Composable agreement by model $\times$ content type.}
\label{fig:scorecard}
\end{minipage}\hfill
\begin{minipage}{0.52\linewidth}\centering
\includegraphics[width=\linewidth]{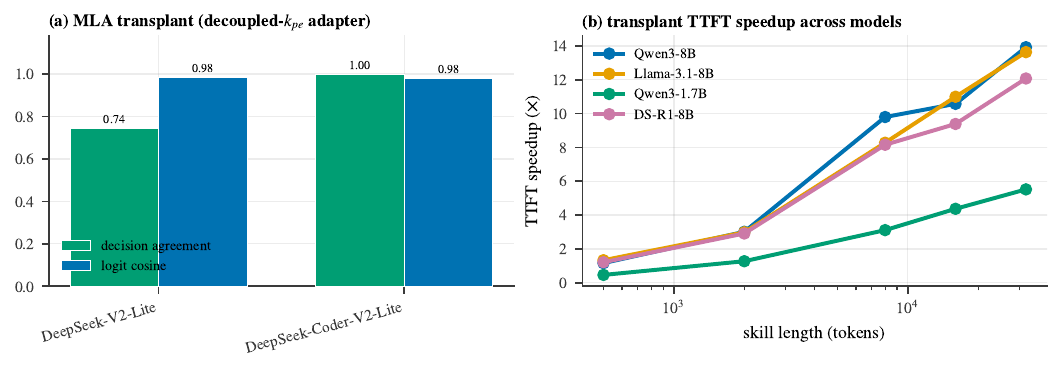}
\caption{MLA adapter fidelity (a) and transplant TTFT speedup across models (b).}
\label{fig:mlascale}
\end{minipage}
\end{figure}

\begin{figure}[tbp]
\centering
\includegraphics[width=\linewidth]{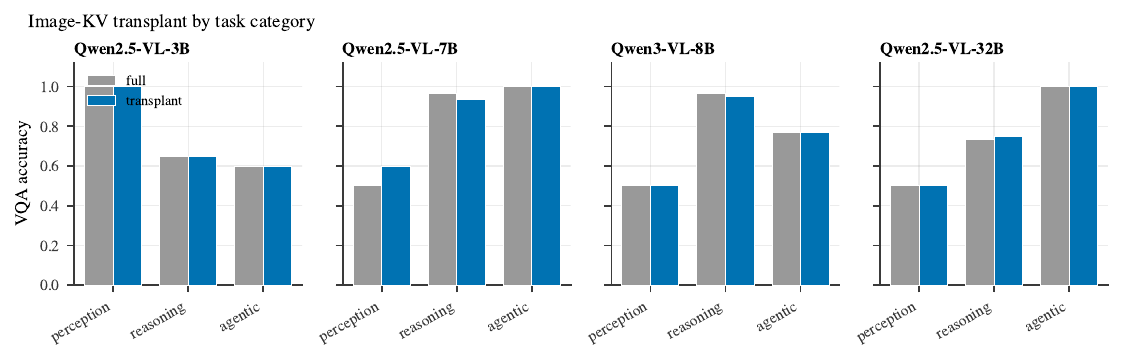}
\caption{Image-KV transplant by task category (perception / reasoning / agentic), full re-encode vs.\
transplant, across four vision-language models. The transplant tracks full accuracy category-by-category.}
\label{fig:mmcat}
\end{figure}

\begin{figure}[tbp]
\centering
\includegraphics[width=\linewidth]{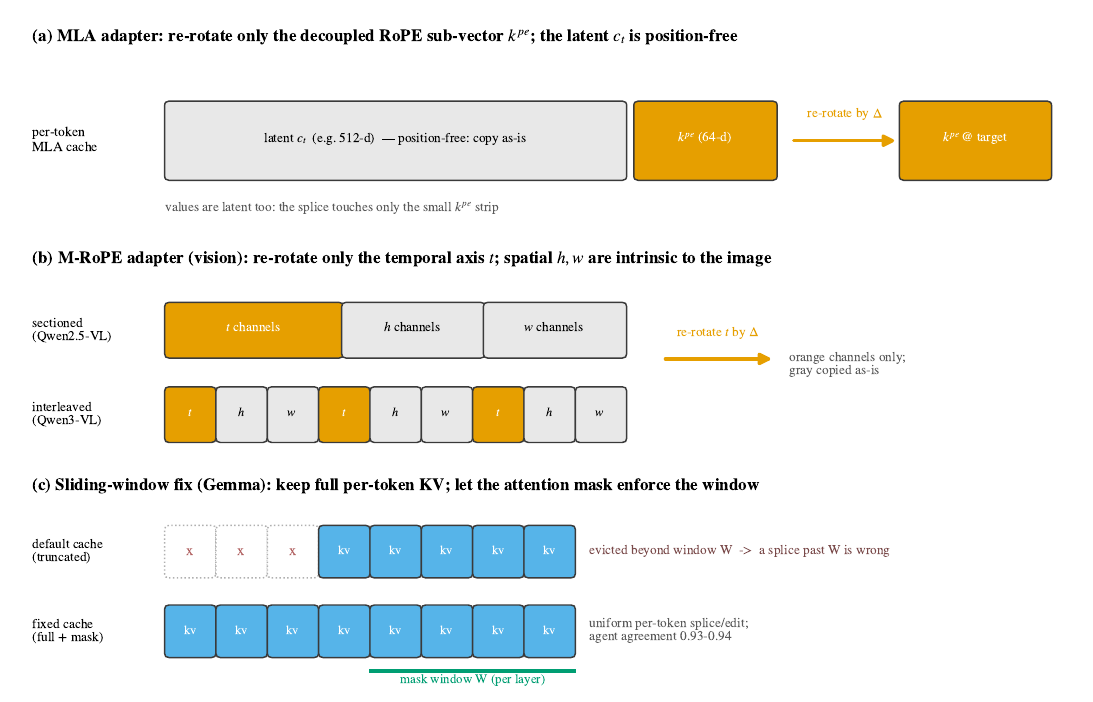}
\caption{\textbf{The attention-variant adapters.} (a) MLA caches a position-free latent $c_t$ plus a small
decoupled-RoPE sub-vector \kpe{}; repositioning re-rotates only \kpe{} and copies the latent as-is. (b)
M-RoPE factors the rotary channels into temporal/height/width axes---sectioned (Qwen2.5-VL) or interleaved
(Qwen3-VL); moving an image within a trajectory re-rotates only the temporal channels, since the spatial
axes are intrinsic to the image. (c) Sliding-window layers default to a window-truncated cache, which
breaks any splice past $W$; keeping the full per-token KV and letting the attention mask enforce the
window restores uniform edit/splice semantics.}
\label{fig:adapters}
\end{figure}

\Cref{fig:scorecard} is the composable scorecard: decision agreement between a transplanted skill and full
recompute, by model and content type (facts in two insertion points, and agentic tool-calling). Standard
attention models are at or near $1.0$; the sliding-window Gemma models are the consistent exception
(addressed in \Cref{sec:reach}). \Cref{fig:mmcat} breaks the multimodal result down by task category, and
\Cref{fig:mlascale} reports the MLA decoupled-\kpe{} adapter fidelity and the transplant TTFT speedup
across models. \Cref{fig:adapters} diagrams the three attention-variant adapters of \Cref{sec:reach}: what
each representation caches, and exactly which slice of it the reposition touches.

\section{Long-horizon robustness and the cross-referential memory test}
\label{app:robustness}

This appendix collects two stress-tests for the leave-stale$+$erratum scheme. The first asks whether
per-edit errors \emph{compound} over a long agent trajectory (they do not); the second asks whether
splitting \emph{cross-referential} facts across independently-precompiled blocks breaks a two-hop chain---the
case \Cref{sec:memory}'s E4 left open (it does, and the mechanism explains why).

\begin{figure}[tbp]
\centering
\includegraphics[width=\linewidth]{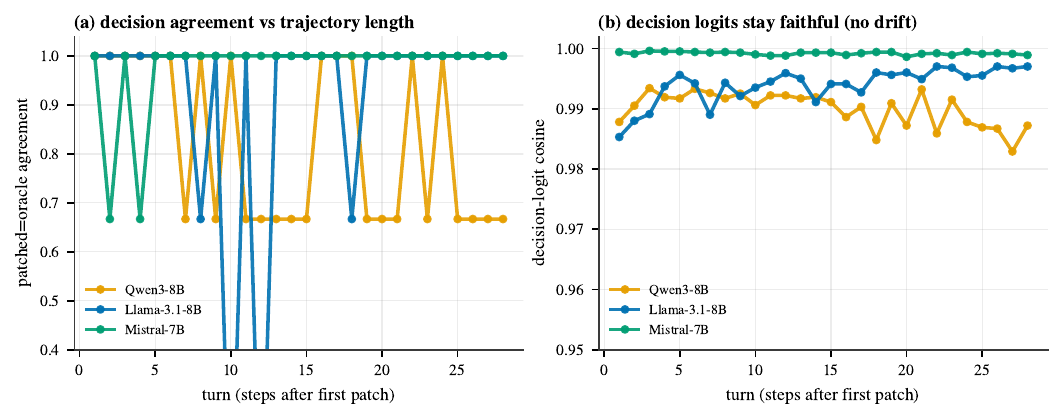}
\caption{\textbf{No compounding error over a long trajectory.} A gated field toggles every turn for $28$
turns; one evolving leave-stale$+$erratum cache vs.\ full reprefill of the identical text. (a) Per-turn
decision agreement stays high with boundary noise (no downward trend). (b) The decision-logit cosine stays
flat at $0.99{+}$---no drift with trajectory length.}
\label{fig:horizon}
\end{figure}

\paragraph{No compounding error over a long trajectory (\Cref{fig:horizon}).} A standing risk for any
leave-stale scheme is that small per-edit errors \emph{compound} over a long agent trajectory. We test it
directly. A single gated field (a clearance level) toggles between a granting and a denying value every turn
over a $28$-turn trajectory; we maintain ONE evolving KV cache---reuse the static prefix forever, apply each
state change as an appended erratum, never recompute the downstream---and compare its per-turn decision to a
\emph{full reprefill of the byte-identical token sequence} (errata included). The only difference between the
two is whether downstream KV was recomputed after each change, so per-turn agreement isolates exactly the
cost of leaving KV stale as a function of trajectory length. Across three families (Qwen3-8B, Llama-3.1-8B,
Mistral-7B) the next-token \emph{decision logits stay faithful}---cosine $0.987$--$0.999$ with a flat
first-third$\to$last-third profile (e.g.\ $0.992\to0.996$ on Llama)---and the agreement-vs-turn slope is
within $\pm0.01$/turn (no systematic decline). Discrete decision agreement is high in aggregate
($0.79$--$0.99$) but noisier than the cosine, because these gated decisions sit near the action boundary
(oracle accuracy $0.52$--$0.82$), where a sub-percent logit difference can flip a discrete choice---the same
boundary sensitivity noted for greedy CoT in \Cref{sec:memory}. The leave-stale$+$erratum cache does not
degrade with trajectory length: there is no compounding drift, only boundary noise.

\paragraph{Cross-referential facts (the \Cref{sec:memory} E4 test).} E4 found splitting \emph{independent}
relevant facts across independently-precompiled blocks decision-lossless. We test the case it left open---a
genuinely \emph{cross-referential} chain. A memory contains a \texttt{DEFINITION} line~$A$ naming which
setting governs the request, whose value~$B$ lives elsewhere; the decision needs the two-hop resolution
$A\!\to\!B$. We lay $A$ and $B$ so they straddle a block boundary and compare a transplant that \emph{splits}
them (boundary between $A$ and $B$) against one that keeps them \emph{colocated} (boundary moved past both),
holding everything else fixed. On Llama-3.1-8B---which performs this two-hop decision at full-recompute
accuracy $0.85$ ($n{=}80$, balanced)---splitting drops agreement with full recompute to $0.46$ versus $0.76$
colocated, a $0.30$ penalty ($25$ vs.\ $1$ discordant instances, McNemar $p{=}8\times10^{-7}$); an
\emph{independent} fact pair split the same way costs only $0.04$ (Mistral-7B). The mechanism explains it:
block~$B$ precompiled in isolation never attended to its referent~$A$, so the chain is not memoized and the
splice cannot restore it. (Qwen3-8B is at chance on the direct two-hop decision---consistent with
\Cref{sec:memory}'s observation that direct gated decisions are at chance for the reasoning-native family
$\le$32B---so it is uninformative here.) The guidance is simple: keep cross-referential facts within one
precompiled block; independent facts may be split freely.

\begin{figure}[tbp]
\centering
\includegraphics[width=0.85\linewidth]{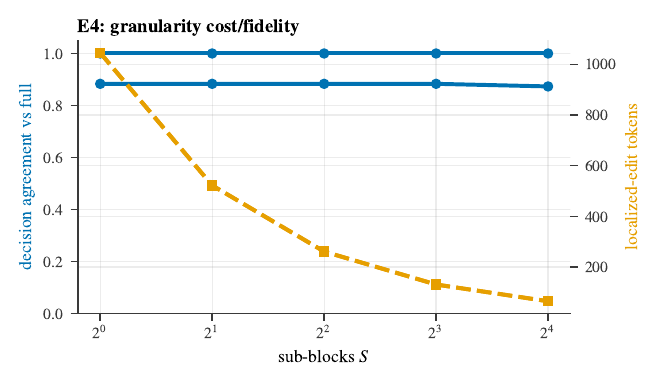}
\caption{\textbf{E4 --- granularity is a free knob} (\Cref{sec:memory}). Splitting memory into $S$
independently-precompiled blocks makes a localized edit $S\times$ cheaper and stays decision-lossless to
$S{=}16$; only genuinely cross-referential facts must be kept in one block (cross-referential test above).}
\label{fig:mem-e4}
\end{figure}

\section{The user-memory agent}
\label{app:memory-agent}
The agent of \Cref{sec:memory} keeps the layout \texttt{[system][trajectory][MEMORY][query]}.
The system prompt is prefilled once; the trajectory grows by cached deltas; the user-memory
chunk is precompiled once in isolation and RoPE-repositioned to float just before the query
each turn (an $O(L_\text{mem})$ re-rotation, no re-prefill) with one boundary token repaired;
a memory change is applied either by recompiling the isolated chunk ($O(L_\text{mem})$, once)
or by appending a salient erratum into the cached trajectory stream (append-only, composes
with prefix caching). Confirmatory protocol: gated decisions whose governing facts live in a
Markdown memory, balanced labels, chain-of-thought as the competent regime (direct one-shot
memory-gated decisions are at chance for $\le$8B models), with cluster-bootstrap CIs
($10^4$, persona-level), TOST equivalence ($\delta{=}0.03$; cosine $\ge0.98$), GEE-logistic
(cluster-robust), McNemar, and BH-FDR. Faithfulness is read against a token-matched
full-reprefill oracle. Pre-registered hypotheses, margins, and the inclusion gate (oracle
accuracy $\ge0.80$) are released with the code.

\section{Online serving and weight-editing methodology}
\label{app:systems}
\paragraph{Online vLLM serving (\Cref{fig:sys}a).} vLLM V1 \texttt{AsyncLLMEngine}, CUDA graphs enabled
(\emph{not} \texttt{enforce\_eager}), continuous batching, automatic prefix caching on, \texttt{bfloat16},
GPU memory utilization $0.85$. Workload: a shared ${\sim}8{,}066$-token policy (real
$\tau^2$-bench retail policy plus neutral padding) with one mutable field; $96$ requests per arm, $64$
output tokens each. Requests arrive as a \emph{Poisson} process at offered rates $\{2,4,8,16\}$ req/s and
unthrottled (saturation). We timestamp first-token and completion per request in the client loop (TTFT,
TPOT, end-to-end) and read each arm's APC hit-rate from the engine's Prometheus counters
(\texttt{vllm:gpu\_prefix\_cache\_hits}, \texttt{\ldots\_queries}). The in-prefix
baseline writes the new field value
early (invalidating downstream APC blocks); the erratum keeps the old prefix and appends the update.
Headline: throughput speedup grows $1.6\times{\to}14.5\times$ as load rises to saturation; $p90$ TTFT
$53$--$398\times$ lower; APC hit-rate $98.5\%$ vs.\ $1.0\%$.

\paragraph{Weight editing (\Cref{tab:weight}).} ROME is implemented from scratch (uncentered key covariance
$C{=}\mathbb{E}[kk^\top]$ at a mid MLP layer over a text sample, an optimized value $v^\star$, and the
closed-form rank-one update of \texttt{down\_proj}) and \emph{validated on the canonical factual edit}
(``the Eiffel Tower is in'' \emph{Paris}$\to$\emph{Rome}, locality preserved) before use, so the baseline
is faithful. LoRA fine-tunes ($r{=}8$, \texttt{q,v,down} projections) on the stale-context decision until it
flips. All methods are evaluated on the same gated decision; cross-request contamination is measured over
$8$ held-out orders that are genuinely still pending (correct $=$ cancel), and collateral over a $10$-item
battery of unrelated single-token gated decisions.

\end{document}